\documentclass[runningheads]{llncs}

\usepackage[year=2026]{eccv}

\usepackage{eccvabbrv}

\usepackage{graphicx}
\usepackage{booktabs}
\usepackage[accsupp]{axessibility} 

\usepackage{bm}
\usepackage{lmodern}
\usepackage{microtype}

\usepackage{nicefrac}
\usepackage{algorithm,algpseudocode}
\usepackage{xcolor}
\usepackage{colortbl}
\usepackage{url}
\usepackage{amssymb,amsmath,mathtools}
\usepackage{multirow}
\usepackage{tabularx}
\usepackage{pifont}
\usepackage{enumitem}

\usepackage{float}        
\usepackage{placeins}     
\usepackage{stfloats}     
\usepackage[font=small,labelfont=bf]{caption}

\def\thetabm{{\bm{\theta }}}

\providecommand{\best}[1]{\textbf{\textcolor[HTML]{D52815}{#1}}}

\definecolor{eccvblue}{rgb}{0.21,0.49,0.74}
\newcommand{\thdr}[1]{\begin{tabular}[c]{@{}c@{}}#1\end{tabular}} 

\usepackage[breaklinks,colorlinks,allcolors=eccvblue]{hyperref}
\usepackage{orcidlink}

\providecommand{\yid}{\mathbf{y}^{\text{id}}}
\providecommand{\xid}{\mathbf{x}^{\text{id}}}
\providecommand{\yood}{\mathbf{y}^{\text{ood}}}

\providecommand{\xpseudo}{\widehat{\mathbf{x}}^{\text{ood}}}
\providecommand{\xinit}{\tilde{\mathbf{x}}^{\text{ood}}}
\providecommand{\Did}{\mathcal{D}_{\text{id}}}
\providecommand{\Dood}{\mathcal{D}_{\text{ood}}}
\providecommand{\Doodsel}{\mathcal{D}_{\text{ood}}^{\text{sel}}}

\providecommand{\Lid}{\mathcal{L}_{\text{id}}}
\providecommand{\Lood}{\mathcal{L}_{\text{ood}}}

\begin{document}

\title{Generative Manifold Distillation:}
\subtitle{Aligning Restoration Trajectories with Natural Image Prior}

\titlerunning{}
\authorrunning{ }

\author{Yuyang Hu\(^{1,2}\thanks{This work was done during an internship at Google.}\,\),  Mojtaba Sahraee-Ardakan\(^1\), Arpit Bansal
\(^1\), Kangfu Mei\(^1\), \\ Chenyang Qi\(^1\),  \text{Peyman Milanfar}\(^1\), \text{Mauricio Delbracio}\(^1\), \\[.15em]
\(^1\)Google,\(^2\)Washington University in St. Louis\\[.1em]
}
\institute{}

\maketitle

\begin{figure}[H]
    \centering
    \includegraphics[width=\linewidth]{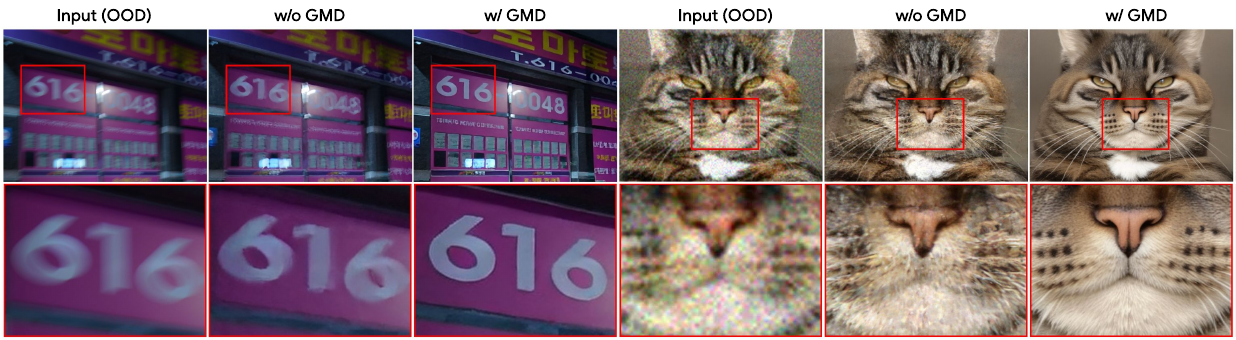}
    \caption{
    \small
    \textbf{Bridging the Domain Gap with GMD.} (Top) Real-world deblurring and (Bottom) 4x super-resolution examples. While the baseline (LDM) fails on out-of-distribution degradations, GMD recovers high-fidelity details without target-domain clean image or architectural modifications.}
    \label{fig:teaser}
\end{figure}

\begin{abstract}
    Pre-trained image restoration models often fail on out-of-distribution (OOD) real-world degradations. Adapting to these domains is challenging as real-world data lacks paired ground truth, and unsupervised methods often require unstable architectural changes. We propose Generative Manifold Distillation (GMD), which reframes domain adaptation as geometric manifold alignment. GMD operates in a strictly unpaired setting, requiring \emph{only low-quality (LQ) target observations}. By leveraging the flow-matching dynamics of a frozen text-to-image foundation model, GMD projects off-manifold restorations onto the natural image manifold to generate high-quality pseudo-targets. To ensure stability, a quality-gated manifold filter rejects off-manifold samples, while source-anchored trajectory regularization prevents error accumulation. Ultimately, GMD distills a powerful generative prior into an efficient restoration network. Experiments demonstrate that GMD seamlessly adapts to new distributions using only LQ inputs, drastically improving perceptual quality with zero architectural modifications or added inference latency.
\end{abstract}


\section{Introduction}
\label{sec:intro}
Image restoration leveraging diffusion models has recently achieved impressive results across tasks like super-resolution~\cite{saharia2022image, li2022srdiff, wang2024exploiting}, deblurring~\cite{whang2022deblurring, ren2023multiscale, chen2024hierarchical}, and inpainting~\cite{lugmayr2022repaint, corneanu2024latentpaint}. These models benefit from powerful learned generative priors, demonstrating high fidelity under in-distribution settings. However, their performance often collapses when applied to real-world images with complex, unknown out-of-distribution (OOD) degradations~\cite{wang2021realesrgan, ren2023multiscale}. Because acquiring paired ground-truth for these OOD samples is practically impossible, researchers face a fundamental challenge: \emph{how can we adapt an efficient restoration model, pre-trained on synthetic data, to a new unlabeled real-world domain?}

Previous approaches to this problem fall into two distinct paradigms: {(1) Traditional Unpaired Domain Adaptation (UDA):} Methods relying on CycleGAN-like frameworks~\cite{wang_unsupervised_sr, lu2019unsupervised, chen2024unsupervised} require intrusive architectural modifications, such as adding domain discriminators or auxiliary feature extractors. These changes are complex to tune and break the seamless deployment of standard pre-trained models.
{(2) Zero-Shot Foundation Models:} Large text-to-image models (e.g., the 12B parameter FLUX~\cite{flux2024}) can act as strong external priors via zero-shot projection techniques (like SDEdit~\cite{sdedit} or RF-Solver~\cite{wang2025taming}). While perceptually impressive, applying a massive foundation model at test time introduces {additional inference latency}. Furthermore, without task-specific finetuning to control them, these strong priors frequently dominate the output, leading to semantic hallucinations and content drift in zero-shot results.

These limitations highlight a clear gap: we need the powerful external generative prior of a large-scale foundation model to overcome severe OOD degradations, but we simultaneously require the strict structural fidelity and blazing-fast inference speed of a standard, unmodified restoration network.

To achieve this, we propose \textbf{Generative Manifold Distillation (GMD)}, a framework that reframes domain adaptation as an elegant \emph{offline distillation} task. GMD completely decouples the heavy generative prior from online inference. Instead of running a foundation model at test time, we treat it as an offline T2I generative prior. In \textbf{Stage 1}, we perform a deterministic, single-pass automated projection to map unlabeled OOD inputs onto the natural image manifold, creating a dataset of high-quality pseudo-targets. Crucially, a strictly quality-gated manifold filter rejects off-manifold samples. In \textbf{Stage 2}, we perform \textbf{source-anchored unsupervised distillation}, fine-tuning the efficient baseline model on these filtered targets while rigidly anchoring it to the original synthetic source data to prevent error accumulation.

\vspace{0.5em}
\noindent\textbf{Contributions.} \\
\textbf{(1) Zero-Latency Prior Distillation:} GMD absorbs the massive external prior of a foundation model entirely offline. This translates the prohibitive computational cost of generative priors into a one-time training setup, yielding significant performance improvement with \emph{zero added inference latency}. \\
\textbf{(2) Mitigating Generative Hallucinations:} By framing adaptation as a dual-task optimization, the fast student model interpolates the target's natural texture manifold while relying on the source data for strict fidelity. This effectively avoids the teacher model's hallucinations.\\
\textbf{(3) Architecture-Agnostic Unsupervised Adaptation:} GMD adapts pre-trained models strictly within the Unpaired Domain Adaptation paradigm. It requires absolutely zero manually labeled target data and zero modifications to the underlying inference architecture.

\begin{table}[t]
\centering
\scriptsize
\setlength{\tabcolsep}{3pt}
\caption{\textbf{Adaptation Paradigm Comparison.} GMD bridges the gap between traditional UDA and zero-shot generative priors, enabling high-quality OOD restoration without architectural changes or high inference latency.}
\label{tab:paradigm_comparison}
\begin{tabular}{l c c c c c}
\toprule
\textbf{Method} & \thdr{\textbf{Modifies Base}\\\textbf{Architecture}} & \thdr{\textbf{Found. Model}\\\textbf{Prior}} & \thdr{\textbf{Test-Time}\\\textbf{Found. Model}} & \thdr{\textbf{Inference}\\\textbf{Latency}} & \thdr{\textbf{Hallucination}\\\textbf{Risk}} \\
\midrule
\textit{Unpaired UDA} & \textcolor{red}{\ding{55}} (Yes) & \textcolor{red}{\ding{55}} (No) & \textcolor{green}{\ding{51}} (No) & \textcolor{green}{\ding{51}} (Fast) & \textcolor{red}{\ding{55}} (High) \\
\textit{Zero-Shot}    & \textcolor{green}{\ding{51}} (No)  & \textcolor{green}{\ding{51}} (Yes) & \textcolor{red}{\ding{55}} (Yes) & \textcolor{red}{\ding{55}} (Slow) & \textcolor{red}{\ding{55}} (High) \\
\midrule
\rowcolor[HTML]{E8F4F8}
\textbf{GMD (Ours)}   & \textcolor{green}{\ding{51}} \textbf{(No)} & \textcolor{green}{\ding{51}} \textbf{(Yes)} & \textcolor{green}{\ding{51}} \textbf{(No)} & \textcolor{green}{\ding{51}} \textbf{(Fast)} & \textcolor{green}{\ding{51}} \textbf{(Low)} \\
\bottomrule
\end{tabular}
\end{table}

\section{Background}
\label{sec:Background}

Image restoration, which aims to recover a high-quality image from its degraded observation, has recently been improved by advances in generative modeling, particularly diffusion-based approaches.

\subsection{Image Restoration with Diffusion Models}
Diffusion models have become powerful tools for conditional image generation and restoration. Conditional variants~\cite{lin2024diffbir, xia2023diffir, li2023diffusion, saharia2021sr3, saharia2022image, wang2024exploiting, whang2022deblurring, ren2023multiscale, saharia2022palette, lugmayr2022repaint, corneanu2024latentpaint, hsiao2024ref, hukernel, Mei2022BiNoisingDT, mei2025mmsr, mei2024latent, mei2024codi} directly learn to sample from the posterior distribution conditioned on the low-quality input. This formulation achieves state-of-the-art performance across diverse restoration tasks, including super-resolution~\cite{lin2024diffbir, xia2023diffir, li2023diffusion, saharia2021sr3, saharia2022image, hukernel}, deblurring~\cite{whang2022deblurring, ren2023multiscale}. However, these models remain sensitive to domain shifts. Trained primarily on synthetic degradations, they often fail to generalize to complex real-world degradations, leading to substantial performance degradation~\cite{ren2023multiscale}.

\subsection{Unsupervised Domain Adaptation for Image Restoration}

A key challenge in real-world image restoration is lacking of large-scale, paired datasets. Unsupervised domain adaptation aims to bridge this gap by adapting a model trained on a labeled source domain to an unlabeled target domain.

One prominent line of work learns restoration mappings from unpaired data, often via CycleGAN-like adversarial frameworks~\cite{yi2017dualgan,zhao2022fcl,zhang2023neural,jiangxin2021learning,ren2020neural,jiang2023uncertainty, dong2021learning, chen2024unsupervised}. {Crucially, while these approaches eliminate the need for strictly paired images, they still fundamentally require a curated set of high-quality (HQ) clean images from the target domain} to establish the desired output distribution. Furthermore, establishing these bidirectional mappings suffers from training instability and relies on complex, task-specific architectures~\cite{chen2024unsupervised, liu2025learning, cheng2025unpaired}.

Another strategy explicitly learns a realistic degradation model~\cite{wolf2021deflow} to synthesize training data, or converts out-of-distribution degradations into in-distribution ones prior to restoration~\cite{pham2024blur2blur,wu2024id}. Alternatively, post-processing methods like DeSRA~\cite{xie2023desra} detect and delete generated artifacts after restoration. However, modeling complex degradations remains challenging, and region-based artifact deletion does not actively adapt the model to the target domain. In contrast, GMD performs a global, offline manifold projection to adapt the model to new OOD domains without test-time post-processing or extra inference latency.

Despite progress, existing UDA methods remain highly task-specific and data-restrictive~\cite{chen2024unsupervised, liu2025learning, cheng2025unpaired}. There is a growing need for a general, model-agnostic approach that can adapt pre-trained restoration models using \emph{strictly low-quality (LQ) target observations}, without requiring any target-domain HQ data or architectural modifications.

\begin{figure}[t]
    \centering
    \includegraphics[width=\linewidth]{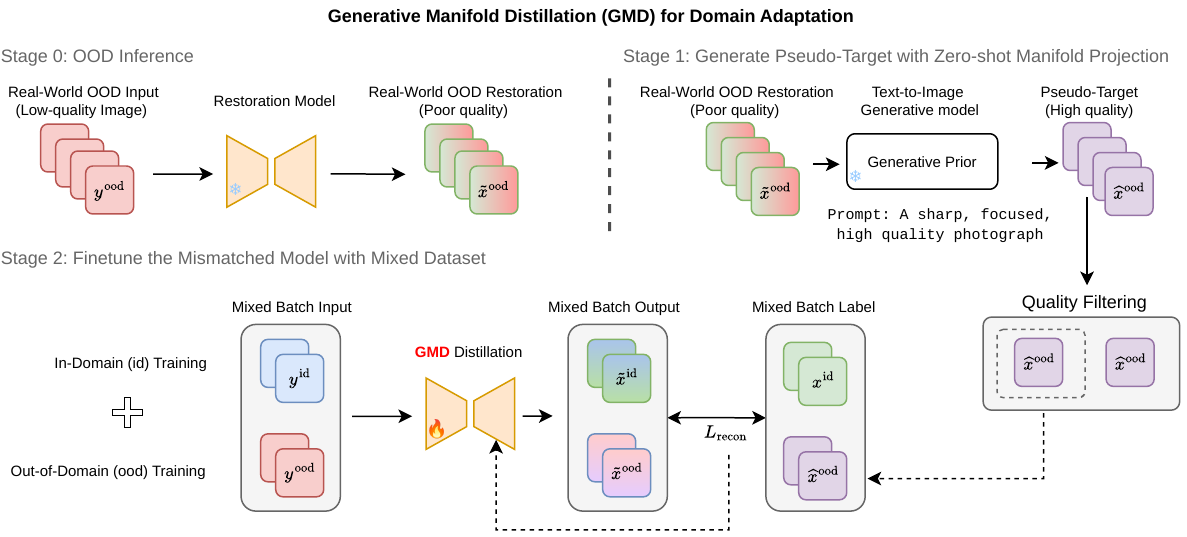}
    \caption{
    \small
    \textbf{Overview of the GMD Framework.} (a) The Challenge: An ID model fails on out-of-distribution $\Dood$ data. Zero-shot foundation models can fix this, but suffer from crippling test-time latency and hallucinations. (b) The GMD Solution: A post-training adaptation process. Stage 1 constructs high-quality pseudo-targets \emph{offline} via zero-shot manifold projection. Stage 2 distills the fast student model on a mix of in-distribution anchors and filtered targets, achieving the generative prior with zero inference latency.
    }
\label{fig:method}
\end{figure}

\subsection{Generative Models as Image Priors}
Large-scale generative models, such as text-to-image diffusion or rectified flow models~\cite{rombach2022high, flux2024, hu2024restoration, hu2025stochastic}, offer powerful priors over natural images. Zero-shot restoration methods leverage these priors by mapping degraded inputs to an intermediate noisy state, followed by guided reconstruction. This is typically achieved either via stochastic perturbation (e.g., SDEdit~\cite{sdedit}) or deterministic forward ODE inversion~\cite{mokady2023null, rout2025semantic, wang2025taming}.

However, deploying these zero-shot strategies is often impractical for high-quality real-world image restoration. They suffer from a severe fidelity-realism trade-off—where the strong, unconstrained prior frequently causes semantic hallucinations and detail loss~\cite{qi2024spire, hukernel}—and their massive parameter counts introduce prohibitive inference costs that make them computationally slow and expensive~\cite{mei2024bigger}.

Instead of deploying a massive foundation model online, GMD fundamentally repurposes it as an offline teacher. By treating the foundation model as an offline manifold projection operator, we distill its massive prior into a fast, dedicated restoration network using only unannotated target data. This entirely bypasses the test-time computational bottleneck and actively regularizes against structural hallucinations.

\section{Method: Generative Manifold Distillation (GMD)}
\label{sec:method_gmd}

The fundamental objective of GMD is to adapt an efficient restoration model $f_{\bm{\theta}}$, pre-trained on a labeled in-distribution (ID) source domain, to a distinct out-of-distribution (OOD) target domain. Crucially, we assume the strictest unsupervised setting for the target domain: we have access \emph{only} to unannotated, low-quality (LQ) degraded observations. Formally, our data regimes are:
\begin{itemize}[leftmargin=2em, topsep=2pt, itemsep=2pt]
    \item \textbf{Source Domain (Paired):} $\Did = \{(\yid_i, \xid_i)\}_{i=1}^{N_{\text{id}}}$
    \item \textbf{Target Domain (LQ-Only):} $\Dood = \{\yood_j\}_{j=1}^{N_{\text{ood}}}$
\end{itemize}

From a representation learning perspective, let $\mathcal{M} \subset \mathbb{R}^d$ represent the low-dimensional manifold of natural, high-fidelity images. A robustly trained ID restorer accurately approximates the projection onto this manifold, mapping $\mathbb{E}_{\mathbb{P}_S}[\mathbf{x}|\mathbf{y}] \in \mathcal{M}$. However, when subjected to real-world OOD data, the model suffers from \textbf{Manifold Drift}. Because the unmodeled OOD degradation shifts the input distribution, the restorer's mapping is pushed into undefined regions of the latent space. Consequently, its outputs exhibit high divergence from the target manifold, characterized by residual artifacts and unnatural textures: $D_{KL}(f_{\bm{\theta}_{\text{id}}}(\mathbb{P}_T(\mathbf{y})) \Vert \mathcal{M}) > \epsilon$. 

To systematically correct this drift relying \emph{only on the LQ inputs $\yood$}, our framework formulates domain adaptation as a manifold alignment process comprising three phases: Initial Off-Manifold Inference, Generative Prior-Guided Orthogonal Projection, and Trajectory-Regularized Distillation.

\subsection{Stage 0: Initial Off-Manifold Inference}
\label{sec:stage0}
For each unlabeled degraded image $\yood \in \Dood$, we first obtain an initial prediction:
\begin{equation}
\xinit = f_{\bm{\theta}_{\text{id}}}(\yood).
\end{equation}
Empirically, $\xinit$ successfully recovers the global structural topology of the scene, but it has been displaced off $\mathcal{M}$. Because the base model lacks the representational capacity for OOD noise profiles, it deposits the image in a low-density ``artifact space'' just outside the boundaries of the natural image manifold. This off-manifold estimate serves as the structural anchor for our subsequent projection.

\subsection{Stage 1: Generative Prior-Guided Orthogonal Projection}
\label{sec:stage1}
To pull the off-manifold estimate $\xinit$ back onto $\mathcal{M}$ without altering its semantic identity, we utilize a frozen large-scale text-to-image generative model as a \textbf{Prior Projection Operator} $\Pi_{\mathcal{G}}: \mathbb{R}^d \to \mathcal{M}$. Let $p_{\text{data}}(\mathbf{x})$ denote the true distribution of natural images, whose high-density support defines the manifold $\mathcal{M}$. The off-manifold input can be modeled as $\xinit = \mathbf{x}_{\text{clean}} + \bm{\epsilon}_{\text{art}}$, where $\bm{\epsilon}_{\text{art}}$ represents unmodeled OOD artifacts pushing the image into a low-probability region ($p_{\text{data}}(\xinit) \approx 0$).

\vspace{0.5em}
\noindent\textbf{(1) Manifold Lifting (Forward ODE):} We first lift $\xinit$ into the generative prior's latent distribution via forward-time Ordinary Differential Equation (ODE) dynamics under null conditioning $\mathbf{c}_\emptyset$:
\begin{equation}
\frac{d\mathbf{x}}{d\tau} = \mathbf{v}(\mathbf{x}(\tau), \tau, \mathbf{c}_\emptyset), \quad \mathbf{x}(0) = \xinit, \quad \tau \in [0,1].
\end{equation}
In the latent space, this forward mapping acts as an artifact-marginalization step. By integrating the deterministic flow to the high-noise state $\mathbf{z} := \mathbf{x}(1)$, the unstructured artifact penalty $\bm{\epsilon}_{\text{art}}$ is asymptotically absorbed into the isotropic noise prior. Simultaneously, the low-frequency spatial coordinates of the scene are encoded into the trajectory's self-attention maps.

\vspace{0.5em}
\noindent\textbf{(2) Orthogonal Projection (Reverse ODE):} We then reconstruct a refined, manifold-aligned image $\xpseudo$ by solving the guided reverse ODE:
\begin{equation}
\xpseudo = \Pi_{\mathcal{G}}(\xinit) = \int_{1}^{0} \mathbf{v}_{\text{cfg}}(\mathbf{x}(\tau), \tau, \mathbf{c}_{\text{prompt}}) d\tau
\label{eq:projection}
\end{equation}
where $\mathbf{v}_{\text{cfg}}$ represents the classifier-free guided velocity field. 

\textbf{Mathematical Interpretation of the Projection:} In generative models, the reverse velocity field $\mathbf{v}_{\text{cfg}}$ inherently points toward high-density regions of the data distribution. Because the degraded input $\xinit$ contains unmodeled OOD artifacts, it initially lies in a low-density region off the natural image manifold $\mathcal{M}$. Integrating the reverse ODE continuously pulls the trajectory toward the high-density regions, mathematically acting as a projection operator that guarantees the output lands on $\mathcal{M}$.

\textbf{Preventing Identity Shift via Attention Injection:} While the reverse ODE guarantees $\xpseudo \in \mathcal{M}$, an unconstrained trajectory can easily project to an arbitrary high-density point, resulting in severe \emph{identity shift} (e.g., hallucinating a completely different valid image on the manifold). To ensure this projection is \emph{orthogonal}—finding the point on $\mathcal{M}$ structurally closest to the input—we employ attention injection~\cite{wang2025taming}. By replacing the self-attention keys and values in the reverse trajectory with those cached from the forward pass, we rigidly lock the spatial layout. This forces the generative flow to shed the off-manifold artifacts while strictly preserving the original semantic identity.

\subsection{Stage 1b: Quality-Gated Manifold Filtering}
\label{sec:filtering}
The generative prior is not perfect; severe OOD degradations can cause the projection to miss the target manifold entirely. To ensure the distillation process is driven by high-confidence signals, we admit a pseudo-pair $(\yood_i, \xpseudo_i)$ into the filtered training set $\Doodsel$ only if the projection successfully reaches a high-density region of $\mathcal{M}$, measured by a reliability threshold $\alpha$:
\begin{equation}
\Doodsel = \left\{ (\yood_i, \xpseudo_i) \mid s(\xpseudo_i) > \alpha \right\}
\end{equation}
where an objective quality metric $s(\cdot)$ serves as a proxy for manifold proximity. By rejecting off-manifold samples that fall below threshold $\alpha$, this quality-gated manifold filter prevents the student restorer from internalizing unnatural hallucinations and erroneous trajectories.

\subsection{Trajectory-Regularized Manifold Distillation}
\label{sec:stage2}
In the final phase, we solve for the updated student parameters $\bm{\theta}$ to internalize the generative prior's orthogonal projection mapping. If we trained $f_{\bm{\theta}}$ exclusively on the target domain pseudo-pairs $\Doodsel$, the model would suffer from manifold collapse—learning to output visually pleasing textures while forgetting the strict deterministic physical mappings required for accurate restoration.

\textbf{Analysis of the Dual-Objective:} We formulate this fine-tuning as a constrained manifold alignment problem:
\begin{equation}
\mathcal{L} = \underbrace{\frac{1}{B_{\text{id}}} \sum_{i=1}^{B_{\text{id}}} \left\lVert f_{\bm{\theta}}(\yid_i) - \xid_i \right\rVert^2}_{\substack{\text{Trajectory Regularization} \\ (\Lid)}} + \lambda \underbrace{\frac{1}{B_{\text{ood}}} \sum_{j=1}^{B_{\text{ood}}} \left\lVert f_{\bm{\theta}}(\yood_j) - \xpseudo_j \right\rVert^2}_{\substack{\text{Manifold Alignment} \\ (\Lood)}}
\label{eq:gmd_loss}
\end{equation}

The parameter $\lambda$ balances perceptual alignment with structural fidelity. Geometrically, $\Lood$ pulls the student's OOD predictions toward the natural image manifold $\mathcal{M}$, while $\Lid$ acts as a structural anchor. By maintaining a high mixing ratio of ID data, we restrict the model to valid physical restorations, preventing error accumulation. This dual objective forces the student to seamlessly interpolate between its physically accurate source knowledge and the target's natural textures, effectively distilling the generative prior into an efficient, single-pass restorer.

\begin{figure}[h]
    \centering
    \includegraphics[width=0.58\linewidth]{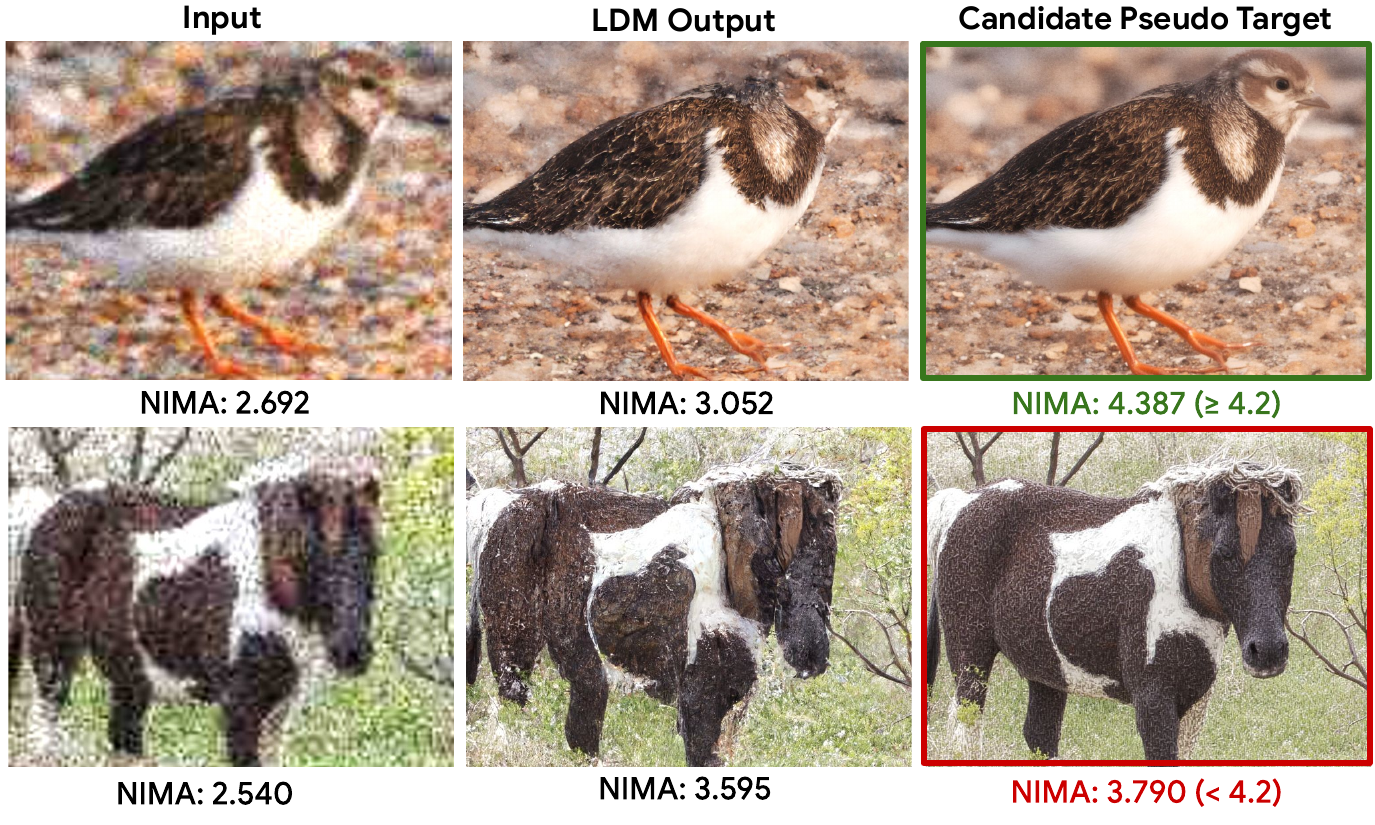}
    \caption{
    \textbf{Selection of projected targets based on image quality assessment.} Top row: A successful selection where the generated target scored above the NIMA threshold ($\ge 4.0$). Bottom row: A rejection case where the projection quality is insufficient.}
    \label{fig:pseudo_selection}
\end{figure}

\FloatBarrier

\section{Experiments}
\label{sec:results}

In this section, we evaluate GMD on synthetic and real-world deblurring and super-resolution. We test its ability to perform domain adaptation using only unlabeled out-of-distribution data, demonstrating that it adapts a pre-trained model to a new domain with zero test-time overhead.

\subsection{Experimental Setup}
\label{sec:exp_setup}

\noindent\textbf{Unpaired Adaptation Data Requirements.} We emphasize that across all experiments, adaptation to the target domain is strictly unsupervised. The models are provided \textbf{only with unlabeled, low-quality (LQ) degraded images} from the target domain's training split. At no point does the model have access to target-domain high-quality (HQ) counterparts, paired data.

We evaluate GMD on the adaptation tasks summarized in Table~\ref{tab:exp_settings}. Our base restoration model, $f_{\thetabm_{id}}$, is a 1.3B parameter Latent Diffusion Model (LDM) based on MMDiT backbone. We test the following adaptation scenarios:
\begin{itemize}[leftmargin=1.2em,itemsep=2pt,topsep=2pt]
    \item \textbf{Deblurring:} Adapting a model trained on GoPro~\cite{nah2017deep} to REDS~\cite{Nah_2019_CVPR_Workshops_REDS} and RealBlur-J~\cite{rim2020real} datasets.
    \item \textbf{Synthetic SR:} Adapting a SR model trained on a \textit{weak} degradation domain (w/ small blur, low noise) to a \textit{strong} domain (larger blur, heavier noise).
    \item \textbf{Real-World SR:} Adapting a SR model trained on synthetic data to the real-world DPED-iPhone dataset~\cite{ignatov2017dslr}.
\end{itemize}
See the supplement for detailed dataset configurations.

\begin{table}[t]
\centering
\scriptsize
\setlength{\tabcolsep}{4pt}
\caption{\textbf{Summary of unsupervised domain adaptation tasks.} We evaluate GMD across four scenarios to test generalization against various distribution shifts. Crucially, the target domain utilizes \textbf{strictly unpaired, low-quality (LQ) images} during adaptation.}
\label{tab:exp_settings}
\begin{tabular}{l l l l}
\toprule
\textbf{Task} & \textbf{Source Domain ($\mathcal{D}_{id}$)} & \textbf{Target Domain ($\mathcal{D}_{ood}$)} & \textbf{Domain Shift} \\
& \textit{(Paired LQ-HQ Data)} & \textit{{(Unpaired LQ Only)}} & \\
\midrule
Deblurring & GoPro~\cite{nah2017deep} & REDS~\cite{Nah_2019_CVPR_Workshops_REDS} & Video-based motion blur. \\
\addlinespace
Deblurring & GoPro~\cite{nah2017deep} & RealBlur-J~\cite{rim2020real} & Real-world low-light \& blur. \\
\addlinespace
SR & Synthetic-SR\textsuperscript{(weak)}~\cite{wang2021realesrgan} & Synthetic-SR\textsuperscript{(strong)}~\cite{wang2021realesrgan} & Degradation Intensity. \\
\addlinespace
SR & Synthetic-SR~\cite{wang2021realesrgan} & DPED-iPhone~\cite{ignatov2017dslr} & Sensor noise \& compression. \\
\bottomrule
\end{tabular}
\end{table}

\begin{figure}[t]
    \centering
    \includegraphics[width=\linewidth]{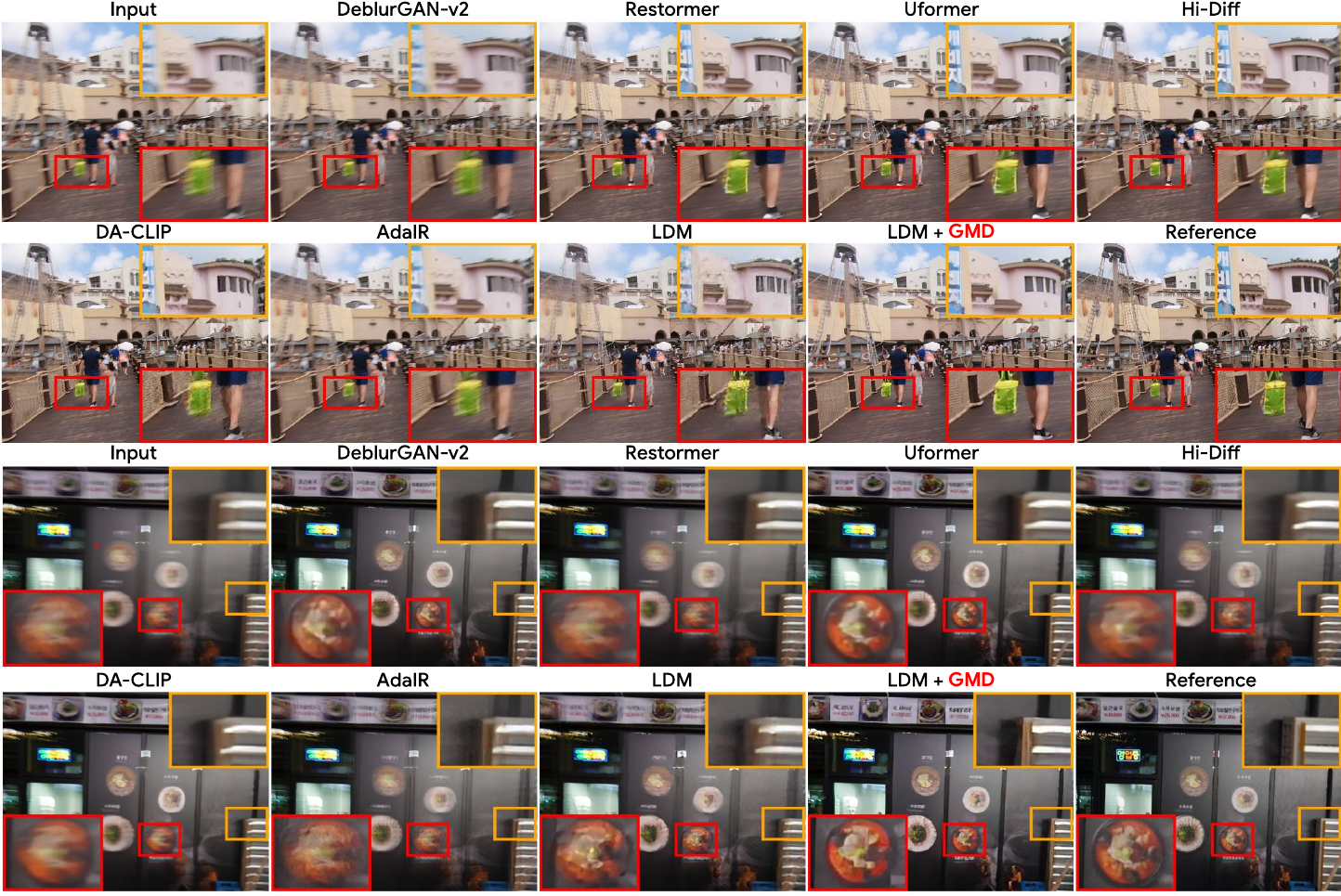}
    \caption{\textbf{Qualitative comparison on the REDS (top) and RealBlur-J (bottom) deblurring dataset.} The GoPro-trained baseline leaves residual blur. GMD successfully adapts to the out-of-distribution domains, producing perceptually superior restorations.}
    \label{fig:deblur_visual}
\end{figure}

\begin{table*}[t]
\centering
\caption{\textbf{Quantitative comparison for real-world deblurring adaptation (GoPro $\rightarrow$ REDS and RealBlur-J).} GMD achieves state-of-the-art performance across all perceptual quality metrics. By vertically stacking the domains, we cleanly present the distortion-perception balance.}
\label{tab:deblur_performance_combined}
\resizebox{\linewidth}{!}{%
\begin{tabular}{l ccccccc}
\toprule
\multirow{2}{*}{\textbf{Method}} & \multicolumn{5}{c}{\textbf{Perceptual Quality}} & \multicolumn{2}{c}{\textbf{Distortion}} \\
\cmidrule(lr){2-6} \cmidrule(lr){7-8}
 & \textbf{LPIPS}$\downarrow$ & \textbf{NIMA}$\uparrow$ & \textbf{MUSIQ}$\uparrow$ & \textbf{FID}$\downarrow$ & \textbf{CLIPIQA}$\uparrow$ & \textbf{PSNR}$\uparrow$ & \textbf{SSIM}$\uparrow$ \\
\midrule
\multicolumn{8}{c}{\cellcolor[HTML]{E8F4F8}\textbf{Evaluation on REDS Dataset}~\cite{Nah_2019_CVPR_Workshops_REDS}} \\
\midrule
DeblurGANv2~\cite{Kupyn_2019_ICCV_Deblurganv2} & 0.190 & 4.350 & 53.17 & 35.77 & 0.313 & 27.07 & 0.805 \\
Restormer~\cite{Zamir2021Restormer} & 0.220 & 4.401 & 53.07 & 36.42 & 0.270 & 26.58 & 0.801 \\
Uformer~\cite{Wang_2022_CVPR_uformer} & 0.197 & 4.315 & 54.24 & 35.11 & 0.283 & 27.20 & 0.825 \\
MPRNet~\cite{Zamir2021MPRNet} & 0.203 & 4.322 & 53.64 & 36.24 & 0.271 & 26.87 & 0.811 \\
Hi-Diff~\cite{chen2024hierarchical} & 0.205 & 4.305 & 54.04 & 39.26 & 0.277 & \best{27.21} & \best{0.831} \\
DA-CLIP~\cite{luo2023controlling} & 0.223 & 4.294 & 48.43 & 42.90 & 0.258 & 25.82 & 0.764 \\
AdaIR~\cite{cuiadair} & 0.285 & 4.137 & 40.97 & 49.67 & 0.227 & 25.77 & 0.774 \\
\midrule
LDM-Deblur (Baseline) & 0.183 & 4.325 & 57.60 & 37.67 & 0.306 & 24.08 & 0.678 \\
\textbf{LDM-Deblur w/ GMD (Ours)} & \best{0.179} & \best{4.460} & \best{63.67} & \best{31.64} & \best{0.404} & 24.35 & 0.682 \\
\midrule
\textit{Fully Supervised*} & 0.169 & 4.578 & 65.04 & 32.49 & 0.462 & 24.51 & 0.686 \\
\midrule
\multicolumn{8}{c}{\cellcolor[HTML]{E8F4F8}\textbf{Evaluation on RealBlur-J Dataset}~\cite{rim2020real}} \\
\midrule
DeblurGANv2~\cite{Kupyn_2019_ICCV_Deblurganv2} & 0.147 & 4.093 & 47.82 & 23.76 & 0.291 & \best{27.20} & \best{0.839} \\
Restormer~\cite{Zamir2021Restormer} & 0.150 & 4.060 & 47.71 & 23.97 & 0.245 & 27.07 & 0.824 \\
Uformer~\cite{Wang_2022_CVPR_uformer} & 0.149 & 4.148 & 49.56 & 23.31 & 0.256 & 27.14 & 0.837 \\
MPRNet~\cite{Zamir2021MPRNet} & 0.149 & 4.088 & 47.77 & 25.05 & 0.239 & 26.99 & 0.833 \\
Hi-Diff~\cite{chen2024hierarchical} & 0.145 & 4.142 & 50.07 & 22.28 & 0.254 & 27.12 & 0.831 \\
DA-CLIP~\cite{luo2023controlling} & 0.239 & 3.859 & 38.96 & 42.26 & 0.236 & 20.53 & 0.680 \\
AdaIR~\cite{cuiadair} & 0.233 & 3.861 & 39.44 & 51.45 & 0.230 & 25.92 & 0.781 \\
\midrule
LDM-Deblur (Baseline) & 0.145 & 4.081 & 50.71 & 25.74 & 0.214 & 26.74 & 0.780 \\
\textbf{LDM-Deblur w/ GMD (Ours)} & \best{0.132} & \best{4.480} & \best{61.33} & \best{20.33} & \best{0.354} & 26.88 & 0.796 \\
\midrule
\textit{Fully Supervised*} & 0.110 & 4.487 & 52.57 & 17.63 & 0.293 & 27.96 & 0.829 \\
\bottomrule
\end{tabular}%
}
\end{table*}

Our base restoration model ($f_{\bm{\theta}_{id}}$) is a 1.3B parameter Latent Diffusion Model (LDM). While 1.3B is large, it remains highly efficient relative to the massive 12B foundation model. It is first pre-trained on its in-distribution dataset for 500K iterations. The GMD adaptation process then operates in two offline phases. Phase 1 (unsupervised pseudo-target generation) takes approximately $\sim$3 hours on 4 TPUv5p chips. Phase 2 (knowledge distillation / fine-tuning) takes approximately $\sim$4 hours on 32 TPUv5p chips. This tractable, one-time offline cost completely eliminates any test-time computational overhead, allowing the baseline model to adapt to new domains with zero added inference latency.

\vspace{4pt}
\noindent\textbf{Teacher Model Details:} The pseudo-target generation (Sec.~\ref{sec:stage1}) uses a 12B pre-trained FLUX.1.dev~\cite{flux2024} Rectified Flow model as the generative prior. We solve the forward (inversion) and reverse (generation) ODEs with an Euler solver with $N=50$ steps, classifier-free guidance ($w=3.5$), and attention injection~\cite{wang2025taming} to improve content preservation.

\vspace{4pt}
\noindent\textbf{Baselines:} We compare GMD against several models: \textit{(1) SOTA open sourced Methods} (for deblurring: DeblurGAN-v2~\cite{Kupyn_2019_ICCV_Deblurganv2}, Restormer~\cite{Zamir2021Restormer}, Uformer~\cite{Wang_2022_CVPR_uformer}, MPRNet~\cite{Zamir2021MPRNet}, Hi-Diff~\cite{chen2024hierarchical}, DA-CLIP~\cite{luo2023controlling}, and AdaIR~\cite{cuiadair}); \textit{(2) in-distribution baseline}, the base LDM $f_{\thetabm_{id}}$ trained \emph{only} on in-distribution data, representing the performance \emph{lower bound} without domain adaptation; and \textit{(3) Fully supervised}, the same base LDM fully-supervised trained on the \emph{target} dataset with groundtruth labels, serving as a practical performance upper bound without domain gap.

\vspace{4pt}

\noindent\textbf{Evaluation metrics:}
Performance is evaluated using distortion metrics (PSNR, SSIM~\cite{wang2004image}) and a suite of perceptual metrics (LPIPS~\cite{zhang2018unreasonable}, FID~\cite{heusel2017gans}, and non-reference metrics (NIMA~\cite{talebi2018nima}, MUSIQ~\cite{ke2021musiq}, NIQE~\cite{saad2012blind}, BRISQUE~\cite{mittal2012no}, CLIPIQA~\cite{wang2023exploring}, MANIQA~\cite{yang2022maniqa}).

\vspace{4pt}
\noindent\textbf{Human evaluation.}
Besides the image quality evaluation metrics, we also conducted user studies to validate our method against leading baselines, as shown in Figure~\ref{fig:human_eval_reds}. To ensure statistical reliability, we used 50 raters for the deblur study and 60 for the super-resolution study. We report win rates with 95\% confidence intervals. Further human evaluation details are in the supplement.

\subsection{Main Results}
\label{sec:main_results}

\begin{figure}[H]
    \centering
    \begin{minipage}[c]{0.58\linewidth}
        \centering
        \includegraphics[width=\linewidth]{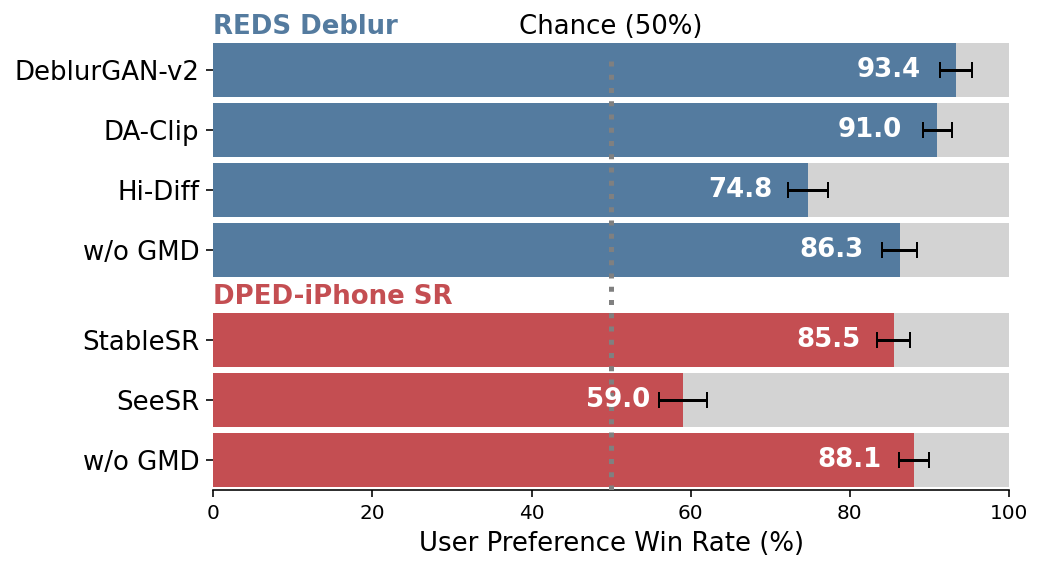}
    \end{minipage}\hfill
    \begin{minipage}[c]{0.38\linewidth}
        \small \textit{``Click on the image that you think is of highest quality (fewer defects, distortions, artifacts, excessive blur, etc.). If both have the same quality, choose the one that appears more natural or pleasing to your eye overall. If both images are equally appealing, click on `Equally Good/Bad'.''}
    \end{minipage}

    \caption{\small \textbf{Human preference study.} (Left) Visual comparisons for deblurring on REDS dataset and super resolution on DPED-iPhone dataset. (Right) The evaluation prompt. GMD consistently outperforms baselines in visual quality. Error bars indicate 95\% CI.}
    \label{fig:human_eval_reds}
\end{figure}

\noindent\textbf{Deblurring adaptation: GoPro $\rightarrow$ REDS / RealBlur-J.}
Table~\ref{tab:deblur_performance_combined} shows that the baseline model trained on GoPro dataset degrades significantly on REDS and RealBlur-J datasets due to domain mismatch. GMD consistently improves the base model across all perceptual quality metrics and achieves the SOTA performance. While distortion-oriented methods like Hi-Diff~\cite{chen2024hierarchical} and DeblurGAN-v2~\cite{Kupyn_2019_ICCV_Deblurganv2} offer strong PSNR/SSIM, GMD delivers superior perceptual realism. As shown in Figures~\ref{fig:deblur_visual}, GMD outputs are consistently sharper and visually closer to the reference than both the unadapted baseline and other SOTA methods. These visual improvements are further validated by a human preference study (Figure~\ref{fig:human_eval_reds}), where GMD is overwhelmingly favored over all the leading baselines.

\vspace{4pt}
\noindent\textbf{Synthetic SR adaptation: Synthetic-SR\textsuperscript{(weak)} $\rightarrow$ Synthetic-SR\textsuperscript{(strong)}.}
To evaluate generalization across degradation intensities, we pre-train a $4\times$ SR model on \textit{weak} degradations (RealESRGAN-style~\cite{wang2021realesrgan} on DIV2K~\cite{div2k} with smaller blur kernels and lower noise levels). We then adapt this model to a \textit{strong} degradation domain (more severe blur and noise) using unlabeled, heavily-degraded images from the Flickr2K~\cite{lim2017edsr} dataset. The adapted model is then evaluated on the strongly-degraded DIV2K test set, which exhibits a clear domain gap from the pre-trained distribution. As shown in Table~\ref{tab:sr_performance_combined}, the adapted model significantly improves performance, increasing MUSIQ by 3.88, PSNR by 0.29dB, and reducing FID by 1.81. This demonstrates GMD's capacity to generalize across domains without target-domain ground truth.

\begin{table*}[t]
\small
\centering
\renewcommand{\arraystretch}{1.1}
\caption{\textbf{Quantitative Results for SR Adaptation.} GMD effectively adapts models to out-of-distribution domains, consistently improving perceptual quality across synthetic (DIV2K) and real-world (DPED) degradations.}
\label{tab:sr_performance_combined}
\resizebox{0.9\linewidth}{!}{%
\begin{tabular}{l ccccc}
\toprule
\multicolumn{6}{c}{\cellcolor[HTML]{E8F4F8}\textbf{4$\times$ SR: Weak $\rightarrow$ Strong (DIV2K)}} \\
\midrule
\textbf{Method} & \textbf{LPIPS}$\downarrow$ & \textbf{MUSIQ}$\uparrow$ & \textbf{FID}$\downarrow$ & \textbf{CLIPIQA}$\uparrow$ & \textbf{PSNR/SSIM}$\uparrow$ \\
\midrule
LDM-SR w/o GMD & 0.386 & 54.41 & 30.11 & 0.442 & 22.03 / .564 \\
\textbf{LDM-SR w/ GMD (Ours)} & \best{0.368} & \best{58.29} & \best{28.30} & \best{0.488} & \best{22.32} / \best{.578} \\
\textit{Fully Supervised*} & 0.233 & 66.73 & 15.06 & 0.602 & 22.80 / .590 \\
\midrule
\multicolumn{6}{c}{\cellcolor[HTML]{E8F4F8}\textbf{2$\times$ SR: DIV2K $\rightarrow$ DPED-iPhone}} \\
\midrule
\textbf{Method} & \textbf{NIQE}$\downarrow$ & \textbf{BRISQUE}$\downarrow$ & \textbf{MANIQA}$\uparrow$ & \textbf{MUSIQ}$\uparrow$ & \textbf{NIMA}$\uparrow$ \\
\midrule
DA-CLIP~\cite{luo2023controlling} & 7.682 & 25.79 & 0.403 & 37.42 & 3.897 \\
StableSR~\cite{wang2024exploiting} & 4.475 & 19.77 & 0.607 & 58.82 & 4.426 \\
SeeSR~\cite{wu2024seesr} & 5.110 & 19.61 & 0.619 & \best{60.19} & 4.430 \\
\midrule
LDM-SR (Baseline) & 5.467 & 19.02 & 0.543 & 49.30 & 4.158 \\
\textbf{LDM-SR w/ GMD (Ours)} & \best{4.300} & \best{16.35} & \best{0.627} & 59.45 & \best{4.452} \\
\bottomrule
\end{tabular}%
}
\end{table*}

\begin{figure}[h]
    \centering
    \includegraphics[width=0.75\linewidth]{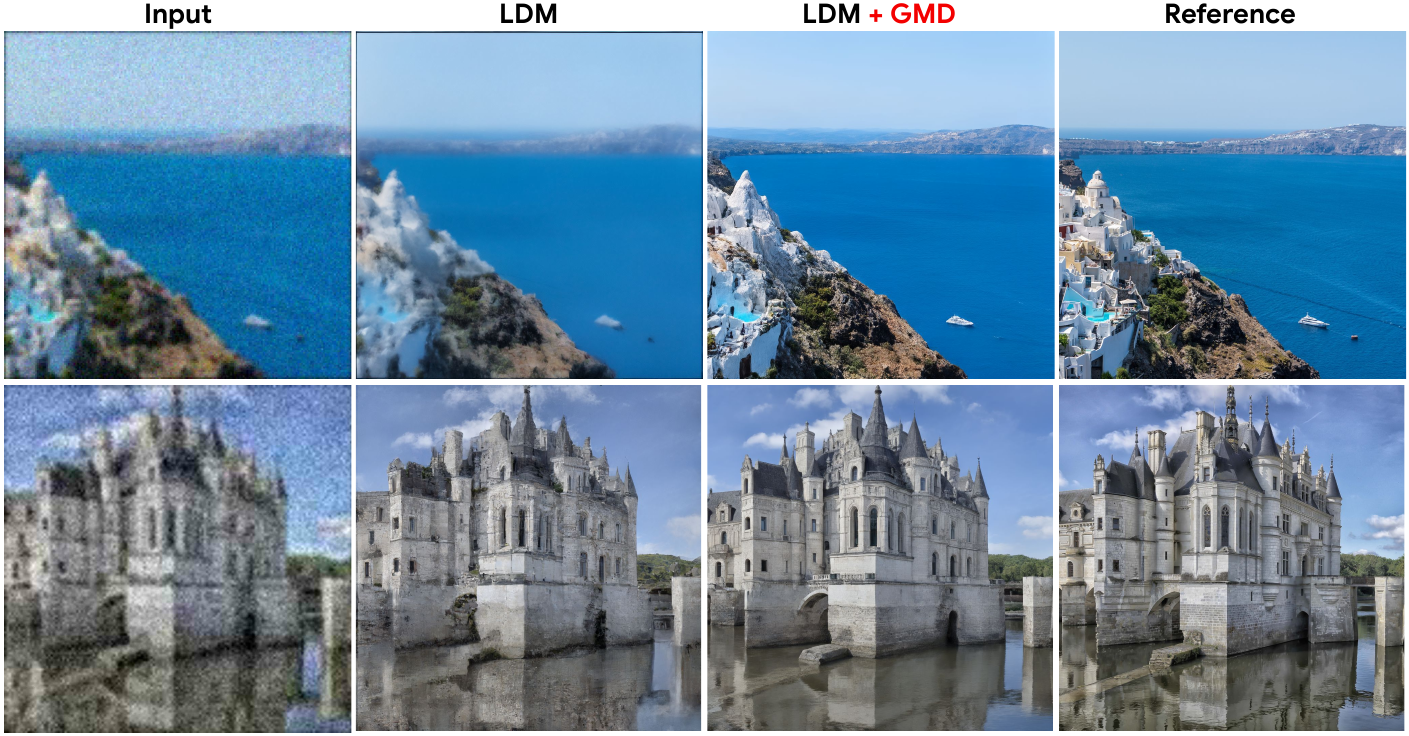}
    \caption{\textbf{Visual comparison on Synthetic Super-Resolution (Weak $\to$ Strong).} The baseline model, pre-trained only on weak degradations, fails to generalize to the heavy noise and blur in the target domain. GMD successfully adapts to the stronger degradation setting, producing clean and sharp restorations.}
    \label{fig:sr_main_visual}
\end{figure}

\begin{figure}[t]
    \centering
    \includegraphics[width=\linewidth]{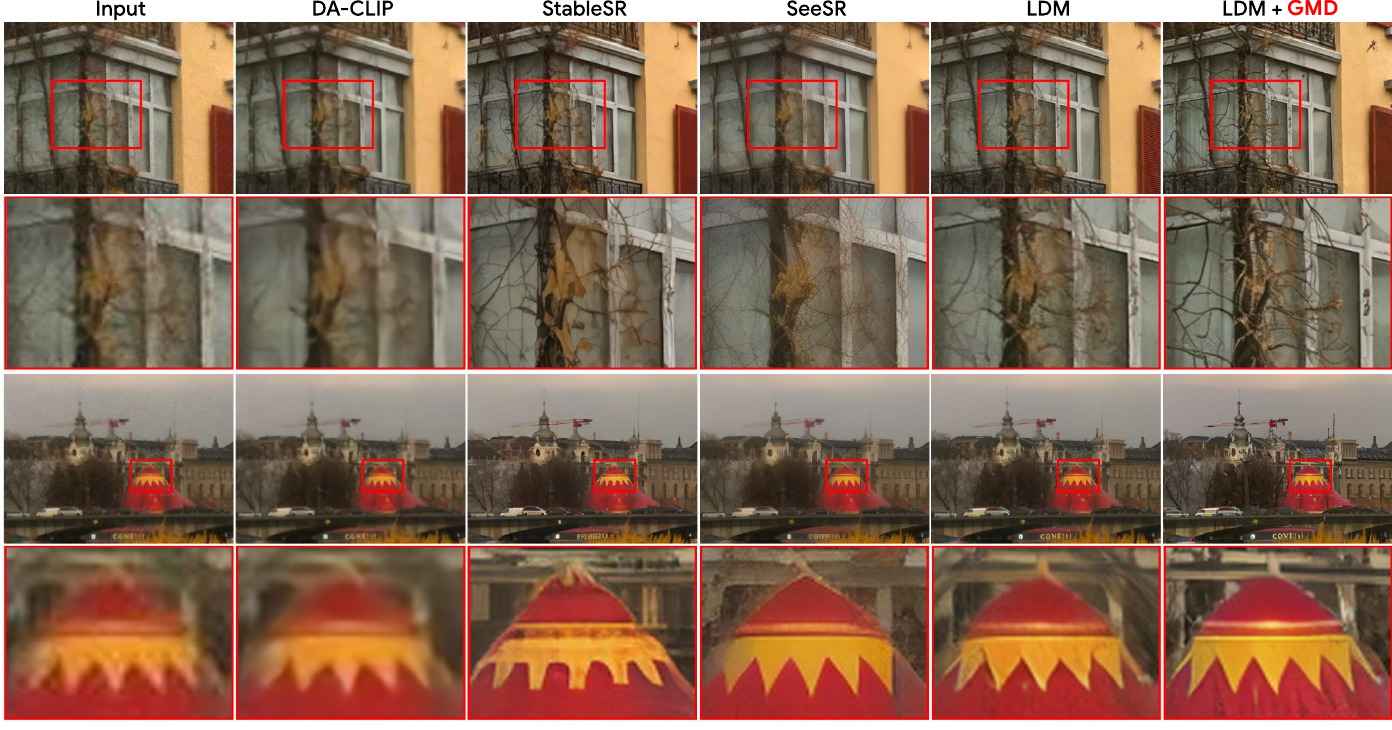}
    \caption{\textbf{Qualitative comparison of real-world SR on the DPED-iPhone dataset.} Compared to state-of-the-art baselines, GMD produces sharper details and more natural textures under real-world degradations without introducing artifacts.}
    \label{fig:iphone_sr}
\end{figure}

\vspace{4pt}
\noindent\textbf{Real-world SR adaptation: Synthetic-SR $\rightarrow$ DPED-iPhone.}
Table~\ref{tab:sr_performance_combined} highlights adaptation from synthetic SR (RealESRGAN-style~\cite{wang2021realesrgan} degradation on DIV2K~\cite{div2k}) to the real-world DPED-iPhone dataset. GMD improves perceptual metrics significantly—e.g., NIQE drops from 5.47 to 4.30, and MANIQA rises from 0.543 to 0.627—surpassing strong baselines such as DA-CLIP~\cite{luo2023controlling}, StableSR~\cite{wang2024exploiting} and SeeSR~\cite{wu2024seesr}. These gains confirm that GMD effectively adapts models pretrained on synthetic degradations to real-world low-resolution data, without paired labels. Visual examples in Figure~\ref{fig:iphone_sr} further demonstrate GMD's advantage: compared to prior methods, our adapted outputs exhibit more natural textures, reduced artifacts and has fewer hallucinations while preserving semantic content. Our human preference study (Figure~\ref{fig:human_eval_reds}) confirms these visual gains, with GMD overwhelmingly outperforming all leading baselines.

\noindent Due to the page limit, additional visual results across all datasets are available in the supplement.

\FloatBarrier

\begin{table*}[!t]
\centering
\scriptsize
\caption{\textbf{Test-time FLUX projection vs. offline distillation (GMD).}}
\label{tab:zero_shot_comparison}
\setlength{\tabcolsep}{2.2pt}
\renewcommand{\arraystretch}{1.03}
\resizebox{\textwidth}{!}{%
\begin{tabular}{l|cccccc|cc}
\toprule
& \multicolumn{6}{c|}{\textbf{Quality}} & \multicolumn{2}{c}{\textbf{Compute}} \\
\textbf{Method} &
LPIPS$\downarrow$ & NIMA$\uparrow$ & MUSIQ$\uparrow$ & FID$\downarrow$ & CLIPIQA$\uparrow$ & PSNR$\uparrow$ / SSIM$\uparrow$ &
\textbf{Params} & \textbf{Latency} \\
\midrule
Baseline (GoPro) & 0.183 & 4.325 & 57.60 & 37.67 & 0.306 & 24.08 / 0.678 & 1.3B & 1.17s \\
+ SDEdit (online) & 0.201 & 4.445 & 63.50 & 40.09 & 0.404 & 20.18 / 0.619 & 1.3B+12B & 1.17s+2.40s \\
+ FLUX.1 Kontext (online) & 0.180 & 4.448 & 63.60 & 32.10 & 0.400 & 21.50 / 0.630 & 1.3B+12B & 1.17s+6.50s \\
+ RF-Solver (online) & 0.186 & 4.439 & 63.47 & 33.57 & 0.404 & 22.15 / 0.643 & 1.3B+12B & 1.17s+4.79s \\
\rowcolor[HTML]{E8F4F8}
\textbf{w/ GMD (offline)} & \best{0.179} & \best{4.460} & \best{63.67} & \best{31.64} & \best{0.404} & \best{24.35} / \best{0.682} & \best{1.3B} & \best{1.17s} \\
\bottomrule
\end{tabular}%
}
\end{table*}

\FloatBarrier

\subsection{Ablation Studies}
\label{sec:ablation_studies}

We analyze: (1) the importance of quality-gated filtering; (2) the impact of the source-anchoring ratio during fine-tuning; and (3) the role of attention injection for zero-shot manifold alignment.

\begin{table}[H]
    \centering
    \scriptsize
    \setlength{\tabcolsep}{6pt}
    \caption{\textbf{Importance of quality-gated filtering.} Filtering targets (Stage 1) is crucial to prevent the student from overfitting to generative artifacts and semantic drift.}
    \label{tab:pseudo_target_filtering}
    \begin{tabular}{l c c c c c}
    \toprule
    \textbf{Variant} & \textbf{PSNR / SSIM}$\uparrow$ & \textbf{LPIPS}$\downarrow$ & \textbf{MUSIQ}$\uparrow$ & \textbf{FID}$\downarrow$ & \textbf{CLIPIQA}$\uparrow$ \\ \midrule
     w/o filter & 23.51 / .652 & 0.293 & 50.17 & 41.26 & 0.272 \\
     \rowcolor[HTML]{E8F4F8} \textbf{w/ filter (Ours)} & \best{24.35} / \best{.682} & \best{0.179} & \best{63.67} & \best{31.64} & \best{0.404} \\
    \bottomrule
    \end{tabular}
\end{table}

\vspace{0.5em}
\noindent\textbf{Effect of target filtering and Content Drift Mitigation.}
High-quality pseudo-supervision is essential for successful adaptation. Table~\ref{tab:pseudo_target_filtering} examines the impact of the NIMA-based quality gate used in Stage 1. Disabling this mechanism causes a dramatic degradation in performance: LPIPS worsens from 0.179 to 0.293, and FID increases from 31.64 to 41.26. Notably, the model trained without filtering performs substantially worse than even the unadapted baseline (Baseline FID: 37.67). This highlights the necessity of quality gating to prevent the model from overfitting to artifacts in the generated data, ensuring that the adaptation process is driven by reliable supervision.

\begin{table}[t]
    \centering
    \scriptsize
    \setlength{\tabcolsep}{10pt}
    \caption{\textbf{Effect of source-anchoring ratio.} Ratio of source ID data to OOD target data during fine-tuning. A high ratio (0.9) provides necessary structural regularization.}
    \label{tab:mixed_rate_comparison}
    \begin{tabular}{l ccc ccc}
    \toprule
    \textbf{Ratio} & \textbf{1.0} & \textbf{0.95} & \textbf{0.9 (Ours)} & \textbf{0.6} & \textbf{0.3} & \textbf{0.0} \\ \midrule
    \textbf{LPIPS}$\downarrow$ & 0.183 & 0.188 & \best{0.179} & 0.187 & 0.198 & 0.217 \\
    \textbf{MUSIQ}$\uparrow$ & 57.60 & 58.10 & \best{63.67} & 62.15 & 62.03 & 52.61 \\
    \textbf{FID}$\downarrow$ & 37.67 & 36.75 & \best{31.64} & 39.09 & 39.73 & 53.56 \\
    \textbf{CLIPIQA}$\uparrow$ & 0.306 & 0.320 & \best{0.404} & 0.379 & 0.375 & 0.287 \\
    \bottomrule
    \end{tabular}
\end{table}

\vspace{0.5em}
\noindent\textbf{Effect of data mixing ratio.}
We study the impact of the mixed-supervision strategy in Table~\ref{tab:mixed_rate_comparison}. The 1.0 ratio (in-distribution-only) is the baseline. Training solely on pseudo-targets (0.0 ratio) performs poorly (FID 53.56), as the model overfits to pseudo-label artifacts and suffers from catastrophic forgetting. The best performance is achieved with a 0.9 ratio (90\% in-distribution data). This confirms the necessity of our mixed-supervision strategy: the in-distribution data provides strong regularization, while the small portion of pseudo-targets guides adaptation.

\begin{figure}[h]
    \centering
    \includegraphics[width=0.9\linewidth]{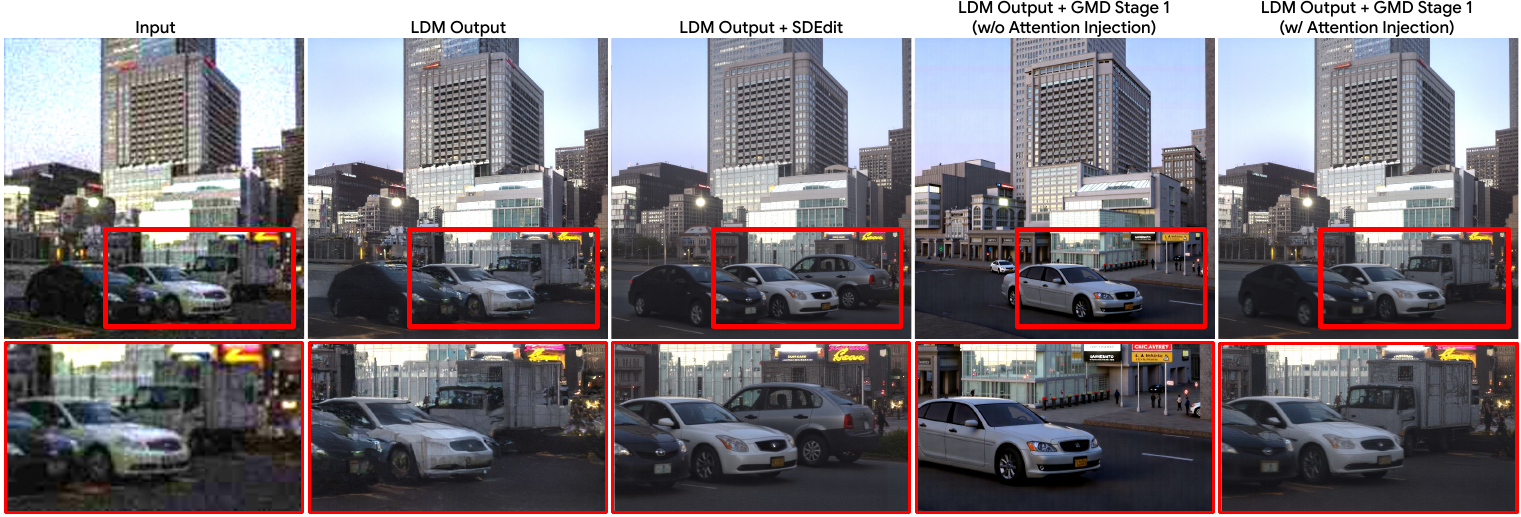}
    \caption{\textbf{Visual comparison of pseudo-target generation strategies (Stage 1).} We evaluate the intrinsic quality of different generation methods. Standard baselines like SDEdit~\cite{sdedit} or the flow matching inversion without Attention Injection help remove artifacts but suffer from severe content drift and structural distortion (e.g., altering vehicle geometry). In contrast, our adopted method with Attention Injection effectively restores details while strictly preserving the original structural layout.}
    \label{fig:pseudo_target_ablation}
\end{figure}

\vspace{0.5em}
\noindent\textbf{Zero-shot Manifold Alignment.} Finally, we assess the visual fidelity of the pseudo-targets generated in Stage 1 (Zero-shot Manifold Alignment). Qualitative comparisons in Figure~\ref{fig:pseudo_target_ablation} demonstrate that methods lacking Attention Injection—specifically SDEdit and the ``No Attention'' pipeline—struggle to maintain structural consistency, often hallucinating new geometries or altering semantic identities. In contrast, RF-Solver leverages attention keys and values from the forward process to guide generation, enforcing strict structural fidelity. The pronounced structural failures of the ``No Attention'' variant render it unsuitable for supervising a restoration model, as it would inevitably bias the network toward learning geometric distortions.

\vspace{0.5em}
\noindent\textbf{Generality across teacher and student architectures.} We verify GMD's framework generality by evaluating different teacher and student capacities. Specifically, we ablate student model flexibility by adapting a lightweight, non-diffusion CNN model (\textbf{SwinIR}) under our framework, and ablate teacher capacity by adapting the model using a $9.2\times$ smaller Stable Diffusion (\textbf{SD}) teacher. Please refer to Section~\ref{sec:supp_generality} in the Supplement for detailed quantitative results, which confirm that GMD generalizes effectively across different restoration backbones and generative prior sources.

\FloatBarrier

\section{Conclusion}
\label{sec:conclusion}

We introduced GMD, an intuitive framework that overcomes the limitations of prior paradigms by transforming unsupervised domain adaptation into an offline generative distillation task. Instead of modifying model architectures, relying on unstable adversarial training, or incurring the prohibitive inference latency of zero-shot foundation models, we leverage a frozen T2I generative model offline to create high-quality pseudo-targets. A source-anchored distillation process ensures the student model learns real-world textures while actively smoothing over the teacher's semantic hallucinations. Extensive experiments demonstrate that GMD effectively bridges the domain gap, achieving state-of-the-art perceptual quality and surpassing the generative fidelity ceiling, all while maintaining zero inference-time overhead.

\FloatBarrier

\section*{Acknowledgements}
The authors would like to thank our colleagues Keren Ye, Viraj Shah and Ashwini Pokle for helpful discussions.

\clearpage
{
    \small
    \bibliographystyle{splncs04}
    \bibliography{main}

\begin{thebibliography}{10}
\providecommand{\url}[1]{\texttt{#1}}
\providecommand{\urlprefix}{URL }
\providecommand{\doi}[1]{https://doi.org/#1}

\bibitem{div2k}
Agustsson, E., Timofte, R.: Ntire 2017 challenge on single image
  super-resolution: Dataset and study. In: CVPRW. pp. 1122--1131. IEEE Computer
  Society (2017)

\bibitem{chen2024unsupervised}
Chen, L., Tian, X., Xiong, S., Lei, Y., Ren, C.: Unsupervised blind image
  deblurring based on self-enhancement. In: Proceedings of the IEEE/CVF
  Conference on Computer Vision and Pattern Recognition. pp. 25691--25700
  (2024)

\bibitem{chen2024hierarchical}
Chen, Z., Zhang, Y., Liu, D., Gu, J., Kong, L., Yuan, X., et~al.: Hierarchical
  integration diffusion model for realistic image deblurring. NIPS  \textbf{36}
  (2024)

\bibitem{cheng2025unpaired}
Cheng, J., Chen, W.T., Lu, X., Yang, M.H.: Unpaired deblurring via decoupled
  diffusion model. arXiv preprint arXiv:2502.01522  (2025)

\bibitem{corneanu2024latentpaint}
Corneanu, C., Gadde, R., Martinez, A.M.: Latentpaint: Image inpainting in
  latent space with diffusion models. In: Proceedings of the IEEE/CVF Winter
  Conference on Applications of Computer Vision. pp. 4334--4343 (2024)

\bibitem{cuiadair}
Cui, Y., Zamir, S.W., Khan, S., Knoll, A., Shah, M., Khan, F.S.: Adair:
  Adaptive all-in-one image restoration via frequency mining and modulation.
  In: The Thirteenth International Conference on Learning Representations
  (2025)

\bibitem{dong2021learning}
Dong, J., Roth, S., Schiele, B.: Learning spatially-variant map models for
  non-blind image deblurring. In: Proceedings of the IEEE/CVF conference on
  computer vision and pattern recognition. pp. 4886--4895 (2021)

\bibitem{heusel2017gans}
Heusel, M., Ramsauer, H., Unterthiner, T., Nessler, B., Hochreiter, S.: Gans
  trained by a two time-scale update rule converge to a local nash equilibrium.
  NIPS  \textbf{30} (2017)

\bibitem{hsiao2024ref}
Hsiao, C.W., Liu, Y.L., Yang, C.K., Kuo, S.P., Jou, K., Chen, C.P.: Ref-ldm: A
  latent diffusion model for reference-based face image restoration. NIPS
  \textbf{37},  74840--74867 (2024)

\bibitem{hu2024restoration}
Hu, Y., Delbracio, M., Milanfar, P., Kamilov, U.: A restoration network as an
  implicit prior. In: International Conference on Learning Representations.
  vol.~2024, pp. 52376--52400 (2024)

\bibitem{hukernel}
Hu, Y., Mei, K., Sahraee-Ardakan, M., Kamilov, U.S., Milanfar, P., Delbracio,
  M.: Kernel density steering: Inference-time scaling via mode seeking for
  image restoration. In: The Thirty-ninth Annual Conference on Neural
  Information Processing Systems (2025)

\bibitem{hu2025stochastic}
Hu, Y., Peng, A., Gan, W., Milanfar, P., Delbracio, M., Kamilov, U.S.:
  Stochastic deep restoration priors for imaging inverse problems. Proceedings
  of Machine Learning Research  \textbf{267},  24621--24652 (2025)

\bibitem{ignatov2017dslr}
Ignatov, A., Kobyshev, N., Timofte, R., Vanhoey, K., Van~Gool, L.: Dslr-quality
  photos on mobile devices with deep convolutional networks. In: Proceedings of
  the IEEE international conference on computer vision. pp. 3277--3285 (2017)

\bibitem{jiang2023uncertainty}
Jiang, R., Han, Y.: Uncertainty-aware variate decomposition for self-supervised
  blind image deblurring. In: Proceedings of the 31st ACM International
  Conference on Multimedia. pp. 252--260 (2023)

\bibitem{jiangxin2021learning}
Jiangxin, D., ROTH, S., SCHIELE, B.: Learning spatially-variant map models for
  non-blind image deblurring. In: CVF Conference on Computer Vision and Pattern
  Recognition, Nashville, USA. pp. 4884--4893 (2021)

\bibitem{ke2021musiq}
Ke, J., Wang, Q., Wang, Y., Milanfar, P., Yang, F.: Musiq: Multi-scale image
  quality transformer. In: ICCV. pp. 5148--5157 (2021)

\bibitem{Kupyn_2019_ICCV_Deblurganv2}
Kupyn, O., Martyniuk, T., Wu, J., Wang, Z.: Deblurgan-v2: Deblurring
  (orders-of-magnitude) faster and better. In: ICCV (Oct 2019)

\bibitem{flux2024}
Labs, B.F.: Flux. \url{https://github.com/black-forest-labs/flux} (2024)

\bibitem{li2022srdiff}
Li, H., Yang, Y., Chang, M., Chen, S., Feng, H., Xu, Z., Li, Q., Chen, Y.:
  Srdiff: Single image super-resolution with diffusion probabilistic models.
  Neurocomputing  \textbf{479},  47--59 (2022)

\bibitem{li2023diffusion}
Li, X., Ren, Y., Jin, X., Lan, C., Wang, X., Zeng, W., Wang, X., Chen, Z.:
  Diffusion models for image restoration and enhancement--a comprehensive
  survey. ARXIV  (2023)

\bibitem{lim2017edsr}
Lim, B., Son, S., Kim, H., Nah, S., Lee, K.M.: Enhanced deep residual networks
  for single image super-resolution. In: CVPRW (2017)

\bibitem{lin2024diffbir}
Lin, X., He, J., Chen, Z., Lyu, Z., Dai, B., Yu, F., Qiao, Y., Ouyang, W.,
  Dong, C.: Diffbir: Toward blind image restoration with generative diffusion
  prior. In: ECCV. pp. 430--448. Springer (2024)

\bibitem{liu2025learning}
Liu, C., Qi, L., Pan, J., Qian, X., Yang, M.H.: Learning deblurring texture
  prior from unpaired data with diffusion model. In: Proceedings of the
  IEEE/CVF International Conference on Computer Vision. pp. 14195--14204 (2025)

\bibitem{lu2019unsupervised}
Lu, B., Chen, J.C., Chellappa, R.: Unsupervised domain-specific deblurring via
  disentangled representations. In: CVPR. pp. 10225--10234 (2019)

\bibitem{lugmayr2022repaint}
Lugmayr, A., Danelljan, M., Romero, A., Yu, F., Timofte, R., Van~Gool, L.:
  Repaint: Inpainting using denoising diffusion probabilistic models. In: CVPR.
  pp. 11461--11471 (2022)

\bibitem{luo2023controlling}
Luo, Z., Gustafsson, F.K., Zhao, Z., Sj{\"o}lund, J., Sch{\"o}n, T.B.:
  Controlling vision-language models for multi-task image restoration. In: ICLR
  (2024), \url{https://openreview.net/forum?id=t3vnnLeajU}

\bibitem{mei2024codi}
Mei, K., Delbracio, M., Talebi, H., Tu, Z., Patel, V.M., Milanfar, P.: Codi:
  Conditional diffusion distillation for higher-fidelity and faster image
  generation. In: Proceedings of the IEEE/CVF Conference on Computer Vision and
  Pattern Recognition. pp. 9048--9058 (2024)

\bibitem{mei2024latent}
Mei, K., Figueroa, L., Lin, Z., Ding, Z., Cohen, S., Patel, V.M.: Latent
  feature-guided diffusion models for shadow removal. In: Proceedings of the
  IEEE/CVF Winter Conference on Applications of Computer Vision. pp. 4313--4322
  (2024)

\bibitem{Mei2022BiNoisingDT}
Mei, K., Nair, N.G., Patel, V.M.: Bi-noising diffusion: Towards conditional
  diffusion models with generative restoration priors. arXiv preprint
  arXiv:2212.07352  (2022)

\bibitem{mei2025mmsr}
Mei, K., Talebi, H., Ardakani, M., Patel, V.M., Milanfar, P., Delbracio, M.:
  The power of context: How multimodality improves image super-resolution. In:
  CVPR (2025)

\bibitem{mei2024bigger}
Mei, K., Tu, Z., Delbracio, M., Talebi, H., Patel, V.M., Milanfar, P.: Bigger
  is not always better: Scaling properties of latent diffusion models. TMLR
  (2024)

\bibitem{sdedit}
Meng, C., He, Y., Song, Y., Song, J., Wu, J., Zhu, J.Y., Ermon, S.: Sdedit:
  Guided image synthesis and editing with stochastic differential equations.
  In: ICLR (2022)

\bibitem{mittal2012no}
Mittal, A., Moorthy, A.K., Bovik, A.C.: No-reference image quality assessment
  in the spatial domain. IEEE Transactions on image processing
  \textbf{21}(12),  4695--4708 (2012)

\bibitem{mokady2023null}
Mokady, R., Hertz, A., Aberman, K., Pritch, Y., Cohen-Or, D.: Null-text
  inversion for editing real images using guided diffusion models. In: CVPR.
  pp. 6038--6047 (2023)

\bibitem{Nah_2019_CVPR_Workshops_REDS}
Nah, S., Baik, S., Hong, S., Moon, G., Son, S., Timofte, R., Lee, K.M.: Ntire
  2019 challenge on video deblurring and super-resolution: Dataset and study.
  In: CVPRW (June 2019)

\bibitem{nah2017deep}
Nah, S., Kim, T.H., Lee, K.M.: Deep multi-scale convolutional neural network
  for dynamic scene deblurring. In: CVPR (July 2017)

\bibitem{pham2024blur2blur}
Pham, B.D., Tran, P., Tran, A., Pham, C., Nguyen, R., Hoai, M.: Blur2blur: Blur
  conversion for unsupervised image deblurring on unknown domains. In:
  Proceedings of the IEEE/CVF Conference on Computer Vision and Pattern
  Recognition. pp. 2804--2813 (2024)

\bibitem{qi2024spire}
Qi, C., Tu, Z., Ye, K., Delbracio, M., Milanfar, P., Chen, Q., Talebi, H.:
  Spire: Semantic prompt-driven image restoration. In: European Conference on
  Computer Vision. pp. 446--464. Springer (2024)

\bibitem{ren2020neural}
Ren, D., Zhang, K., Wang, Q., Hu, Q., Zuo, W.: Neural blind deconvolution using
  deep priors. In: Proceedings of the IEEE/CVF conference on computer vision
  and pattern recognition. pp. 3341--3350 (2020)

\bibitem{ren2023multiscale}
Ren, M., Delbracio, M., Talebi, H., Gerig, G., Milanfar, P.: Multiscale
  structure guided diffusion for image deblurring. In: ICCV. pp. 10721--10733
  (2023)

\bibitem{rim2020real}
Rim, J., Lee, H., Won, J., Cho, S.: Real-world blur dataset for learning and
  benchmarking deblurring algorithms. In: ECCV. pp. 184--201. Springer (2020)

\bibitem{rombach2022high}
Rombach, R., Blattmann, A., Lorenz, D., Esser, P., Ommer, B.: High-resolution
  image synthesis with latent diffusion models. In: CVPR. pp. 10684--10695
  (2022)

\bibitem{rout2025semantic}
Rout, L., Chen, Y., Ruiz, N., Caramanis, C., Shakkottai, S., Chu, W.S.:
  Semantic image inversion and editing using rectified stochastic differential
  equations. In: ICLR (2025), \url{https://openreview.net/forum?id=Hu0FSOSEyS}

\bibitem{saad2012blind}
Saad, M.A., Bovik, A.C.: Blind quality assessment of videos using a model of
  natural scene statistics and motion coherency. In: 2012 Conference Record of
  the Forty Sixth Asilomar Conference on Signals, Systems and Computers
  (ASILOMAR). pp. 332--336. IEEE (2012)

\bibitem{saharia2022palette}
Saharia, C., Chan, W., Chang, H., Lee, C., Ho, J., Salimans, T., Fleet, D.,
  Norouzi, M.: Palette: Image-to-image diffusion models. In: ACM SIGGRAPH 2022
  Conference Proceedings. pp. 1--10 (2022)

\bibitem{saharia2022image}
Saharia, C., Ho, J., Chan, W., Salimans, T., Fleet, D.J., Norouzi, M.: Image
  super-resolution via iterative refinement. PAMI  \textbf{45}(4),  4713--4726
  (2022)

\bibitem{saharia2021sr3}
Saharia, C., Ho, J., Chan, W., Salimans, T., Fleet, D.J., Norouzi, M.: Image
  super-resolution via iterative refinement. PAMI  \textbf{45}(4),  4713--4726
  (2022)

\bibitem{talebi2018nima}
Talebi, H., Milanfar, P.: Nima: Neural image assessment. TIP  \textbf{27}(8),
  3998--4011 (2018)

\bibitem{wang2025taming}
Wang, J., Pu, J., Qi, Z., Guo, J., Ma, Y., Huang, N., Chen, Y., Li, X., Shan,
  Y.: Taming rectified flow for inversion and editing. In: ICML (2025),
  \url{https://openreview.net/forum?id=uDreZphNky}

\bibitem{wang2023exploring}
Wang, J., Chan, K.C., Loy, C.C.: Exploring clip for assessing the look and feel
  of images. In: AAAI. pp. 2555--2563 (2023)

\bibitem{wang2024exploiting}
Wang, J., Yue, Z., Zhou, S., Chan, K.C., Loy, C.C.: Exploiting diffusion prior
  for real-world image super-resolution. IJCV  \textbf{132}(12),  5929--5949
  (2024)

\bibitem{wang_unsupervised_sr}
Wang, W., Zhang, H., Yuan, Z., Wang, C.: Unsupervised real-world
  super-resolution: A domain adaptation perspective. In: ICCV. pp. 4298--4307
  (2021). \doi{10.1109/ICCV48922.2021.00428}

\bibitem{wang2021realesrgan}
Wang, X., Xie, L., Dong, C., Shan, Y.: Real-esrgan: Training real-world blind
  super-resolution with pure synthetic data. In: ICCV. pp. 1905--1914 (2021)

\bibitem{Wang_2022_CVPR_uformer}
Wang, Z., Cun, X., Bao, J., Zhou, W., Liu, J., Li, H.: Uformer: A general
  u-shaped transformer for image restoration. In: CVPR. pp. 17683--17693 (June
  2022)

\bibitem{wang2004image}
Wang, Z., Bovik, A.C., Sheikh, H.R., Simoncelli, E.P.: Image quality
  assessment: from error visibility to structural similarity. TIP
  \textbf{13}(4),  600--612 (2004)

\bibitem{whang2022deblurring}
Whang, J., Delbracio, M., Talebi, H., Saharia, C., Dimakis, A.G., Milanfar, P.:
  Deblurring via stochastic refinement. In: CVPR. pp. 16293--16303 (2022)

\bibitem{wolf2021deflow}
Wolf, V., Lugmayr, A., Danelljan, M., Van~Gool, L., Timofte, R.: Deflow:
  Learning complex image degradations from unpaired data with conditional
  flows. In: CVPR. pp. 94--103 (2021)

\bibitem{wu2024id}
Wu, J.H., Tsai, F.J., Peng, Y.T., Tsai, C.C., Lin, C.W., Lin, Y.Y.: Id-blau:
  Image deblurring by implicit diffusion-based reblurring augmentation. In:
  Proceedings of the IEEE/CVF Conference on Computer Vision and Pattern
  Recognition. pp. 25847--25856 (2024)

\bibitem{wu2024seesr}
Wu, R., Yang, T., Sun, L., Zhang, Z., Li, S., Zhang, L.: Seesr: Towards
  semantics-aware real-world image super-resolution. In: CVPR. pp. 25456--25467
  (2024)

\bibitem{xia2023diffir}
Xia, B., Zhang, Y., Wang, S., Wang, Y., Wu, X., Tian, Y., Yang, W., Van~Gool,
  L.: Diffir: Efficient diffusion model for image restoration. In: ICCV. pp.
  13095--13105 (2023)

\bibitem{xie2023desra}
Xie, L., Wang, X., Chen, X., Li, G., Shan, Y., Zhou, J., Dong, C.: Desra:
  detect and delete the artifacts of gan-based real-world super-resolution
  models. In: Proceedings of the 40th International Conference on Machine
  Learning. pp. 38204--38226 (2023)

\bibitem{yang2022maniqa}
Yang, S., Wu, T., Shi, S., Lao, S., Gong, Y., Cao, M., Wang, J., Yang, Y.:
  Maniqa: Multi-dimension attention network for no-reference image quality
  assessment. In: CVPR. pp. 1191--1200 (2022)

\bibitem{yi2017dualgan}
Yi, Z., Zhang, H., Tan, P., Gong, M.: Dualgan: Unsupervised dual learning for
  image-to-image translation. In: Proceedings of the IEEE international
  conference on computer vision. pp. 2849--2857 (2017)

\bibitem{Zamir2021Restormer}
Zamir, S.W., Arora, A., Khan, S., Hayat, Khan, F.S., Yang, M.H.: Restormer:
  Efficient transformer for high-resolution image restoration. In: CVPR (2022)

\bibitem{Zamir2021MPRNet}
Zamir, S.W., Arora, A., Khan, S., Hayat, Khan, F.S., Yang, M.H., Shao, L.:
  Multi-stage progressive image restoration. In: CVPR (2021)

\bibitem{zhang2018unreasonable}
Zhang, R., Isola, P., Efros, A.A., Shechtman, E., Wang, O.: The unreasonable
  effectiveness of deep features as a perceptual metric. In: CVPR. pp. 586--595
  (2018)

\bibitem{zhang2023neural}
Zhang, Y., Wang, C., Tao, D.: Neural maximum a posteriori estimation on
  unpaired data for motion deblurring. IEEE Transactions on Pattern Analysis
  and Machine Intelligence  (2023)

\bibitem{zhao2022fcl}
Zhao, S., Zhang, Z., Hong, R., Xu, M., Yang, Y., Wang, M.: Fcl-gan: A
  lightweight and real-time baseline for unsupervised blind image deblurring.
  In: Proceedings of the 30th ACM International Conference on Multimedia. pp.
  6220--6229 (2022)

\end{thebibliography}
}

\clearpage
\section{Supplement}
\label{sec:supplement}

This supplementary material provides detailed insights into the experimental setup and additional results that complement the main manuscript. The content is organized as follows:

\begin{itemize}
    \item \textbf{Section~\ref{sec:supp_impl}: Dataset and Implementation Details.} We provide comprehensive descriptions of the dataset configurations, degradation parameters, and specific implementation hyperparameters used in our experiments.
    \item \textbf{Section~\ref{sec:supp_eval}: Evaluation Protocols.} We detail the specific criteria and methodology employed for the human evaluation studies and quantitative metrics.
    \item \textbf{Section~\ref{sec:supp_ablation}: Methodological Analysis and Ablation Studies.} We present statistical analysis of the target quality filtering (pass vs. fail rates). Additionally, we provide an in-depth comparison of generation strategies (specifically analyzing RF-Solver with and without attention injection versus SDEdit) and their impact on downstream adaptation performance.
    \item \textbf{Section~\ref{sec:supp_generality}: Generality and Additional Baseline Comparisons.} We provide additional quantitative comparisons on REDS and DPED-iPhone to demonstrate framework generality across different student/teacher capacities (SwinIR CNN student and SD 1.3B teacher) and evaluate GMD against recent fine-tuned SOTA baselines.
    \item \textbf{Section~\ref{sec:supp_visuals}: Additional Visual Results.} We provide an extensive gallery gallery of qualitative comparisons across diverse scenarios to further substantiate the efficacy of GMD.
\end{itemize}

\subsection{Dataset and Implementation Details}
\label{sec:supp_impl}

\subsubsection{Dataset Settings}

\begin{table}[H]
\centering
\footnotesize
\caption{\textbf{Summary of Dataset Settings.} We evaluate GMD across four distinct domain adaptation scenarios. Note that the \textbf{OOD Target} datasets consist solely of unlabeled, low-quality (LQ) inputs.}
\label{tab:dataset_settings}
\resizebox{\linewidth}{!}{%
\begin{tabular}{lllc}
\toprule
\textbf{Task} & \textbf{ID Source} (Pre-train) & \textbf{OOD Target} (Adapt) & \textbf{Test Set} \\
\midrule
Deblur (REDS) & GoPro (2,103 pairs) & REDS Train (First 6k) & REDS (300) \\
Deblur (RealBlur) & GoPro (2,103 pairs) & RealBlur-J Train (3,758) & RealBlur-J (980) \\
Syn. SR (Wk$\to$Str) & DIV2K Weak (30k pairs) & Flickr2K Strong (First 6k) & DIV2K Val (3k) \\
Real SR (Syn$\to$Real) & DIV2K Syn. (30k pairs) & DPED-iPhone (5,614) & DPED (113) \\
\bottomrule
\end{tabular}%
}
\end{table}

Our evaluation encompasses three primary domain adaptation tasks: Deblurring, Synthetic Super-Resolution, and Real-World Super-Resolution. For all experiments, we enforce a strict separation between the adaptation and evaluation data. The unlabeled images used for GMD adaptation are drawn exclusively from the \textit{training} splits of the target datasets, while the \textit{test} splits are held out entirely and used solely for final evaluation.

\vspace{4pt}
\noindent\textbf{Deblurring (REDS).} We use the GoPro dataset (2,103 pairs) as the in-distribution (ID) source for pre-training. For adaptation, we utilize the REDS training set as the unlabeled out-of-distribution (OOD) data, specifically selecting the first 6,000 low-quality (LQ) images. Performance is evaluated on the REDS test split (300 images), following the protocol in~\cite{ren2023multiscale}.

\vspace{4pt}
\noindent\textbf{Deblurring (RealBlur-J).} Similarly, we use GoPro as the source domain. For adaptation, we use the entire RealBlur-J training set (3,758 unlabeled LQ images). Evaluation is performed on the official RealBlur-J test split (980 images).

\vspace{4pt}
\begin{table}[t]
    \centering
    \begin{minipage}[t]{0.48\textwidth}
        \centering
        \scriptsize
        \caption{\textbf{Synthetic Degradation Parameters.} Comparison of the \textit{Weak} (ID) and \textit{Strong} (OOD) degradation.}
        \label{tab:syn_sr_params}
        \resizebox{\linewidth}{!}{%
        \begin{tabular}{lcc}
        \toprule
        Parameter & Weak (ID) & Strong (OOD) \\
        \midrule
        Blur Kernel Sizes & \{7, 9, 11\} & \{17, 19, 21\} \\
        Gaussian Noise ($\sigma$) & $[1, 20] / 255$ & $[20, 30] / 255$ \\
        Poisson Noise Scale & $[0.05, 2.0]$ & $[0.15, 3.0]$ \\
        Sinc Filter Kernel & \{7, 9, 11\} & \{17, 19, 21\} \\
        \bottomrule
        \end{tabular}%
        }
    \end{minipage}\hfill
    \begin{minipage}[t]{0.48\textwidth}
        \centering
        \scriptsize
        \caption{\textbf{Target Quality Filtering Statistics.} Candidates are filtered based on a NIMA threshold ($\alpha=4.2$).}
        \label{tab:filter_stats}
        \resizebox{\linewidth}{!}{%
        \begin{tabular}{lccc}
        \toprule
        Dataset & Total Gen. & Thresh. ($\alpha$) & Pass Rate \\
        \midrule
        REDS & 6,000 & 4.2 & 69.1\% \\
        RealBlur-J & 3,758 & 4.2 & 74.0\% \\
        Flickr2K & 6,000 & 4.2 & 86.4\% \\
        DPED & 5,614 & 4.2 & 84.3\% \\
        \bottomrule
        \end{tabular}%
        }
    \end{minipage}
\end{table}

\noindent\textbf{Synthetic Super-Resolution.}
We focus on the $4\times$ super-resolution task with an output resolution of $1024 \times 1024$. The ID model is trained on DIV2K (approx. 30,000 cropped pairs) using a \textit{weak} degradation model. We then adapt this model to a \textit{strong} degradation domain using the Flickr2K dataset (first 6,000 LQ images). Both settings employ a high-order degradation pipeline inspired by Real-ESRGAN, involving randomized sequences of blur, resizing, noise injection, and JPEG compression. The \textit{strong} domain (OOD) represents a significant distribution shift, characterized by substantially larger blur kernels and higher noise intensities compared to the source domain. Specific parameter differences are detailed in Table~\ref{tab:syn_sr_params}.

\vspace{4pt}
\noindent\textbf{Real-World Super-Resolution.} We evaluate on the $2\times$ super-resolution task with an output resolution of $512 \times 512$. We first pre-train the model on DIV2K using a standard synthetic RealESRGAN degradation pipeline. We then adapt it to the real-world domain using unlabeled images from the DPED-iPhone training set (5,614 LQ images). Testing is conducted on the standard 113 test images from DPED-iPhone; following standard protocol, we extract and evaluate the center $256 \times 256$ crop of each test image.

\subsubsection{Implementation Details}

\paragraph{Restoration Model.}
Our base model is a computationally tractable 1.3B parameter Latent Diffusion Model (LDM) with an MMDiT backbone. It is pre-trained on the source domain for 500K iterations with a learning rate of $10^{-4}$. During the GMD adaptation stage, we fine-tune for 20K iterations with a reduced learning rate of $5\times 10^{-5}$. We use a batch size of 32 distributed across 32 TPUv5p chips.

\subsection{Evaluation Protocols}
\label{sec:supp_eval}

To comprehensively assess restoration quality, particularly regarding perceptual realism, we conducted large-scale human evaluation studies for both deblurring and super-resolution tasks. This section details the criteria and protocols used.

\vspace{4pt}
\noindent\textbf{Human Evaluation Setup.} We utilized a pairwise preference protocol to systematically compare our proposed method against leading baselines. For each comparison, raters were presented with two anonymized side-by-side images. The raters were provided with the following specific instructions:
\begin{quote}
\textit{``Click on the image that you think is of highest quality (fewer defects, distortions, artifacts, excessive blur, etc.). If both have the same quality, choose the one that is more appealing to you (more interesting, better composition, etc.). If both images are equally appealing, click on `Equally Good/Bad'.''}
\end{quote}

\vspace{4pt}
\noindent\textbf{Data Collection and Quality Control.} The evaluations were conducted on a crowdsourcing platform using a diverse pool of raters to minimize subjective bias. The deblurring task study included 50 unique raters, while the super-resolution task included 60 unique raters.

\subsection{Methodological Analysis and Ablation Studies}
\label{sec:supp_ablation}

\noindent\textbf{Target Quality Filtering and Content Drift Mitigation.}
A critical component of GMD is the quality-gated selection of targets. Generative zero-shot restoration methods (such as SDEdit or DDIM inversion) are not error-free; they are susceptible to semantic hallucination (e.g., generating visually sharp but semantically incorrect text characters or unnatural textures in highly degraded regions). When the input OOD degradation is exceptionally severe, the geometric anchoring in Stage 1 may fail to perfectly constrain the FLUX model, yielding outputs characterized by artifacts or structural collapse. Including these failed restorations in the training set would introduce noise and encourage the model to learn geometric distortions. To mitigate this, our filtering mechanism serves as a crucial outlier rejection step. Because semantically drifted images frequently exhibit localized artifacts, structural incoherence, or unnatural boundary transitions, they predominantly fail to meet the NIMA quality threshold and are systematically excluded from the distillation set. This ensures the student model safely learns texture realism without internalizing semantic hallucinations.

\begin{figure}[t]
    \centering
    \includegraphics[width=0.55\linewidth]{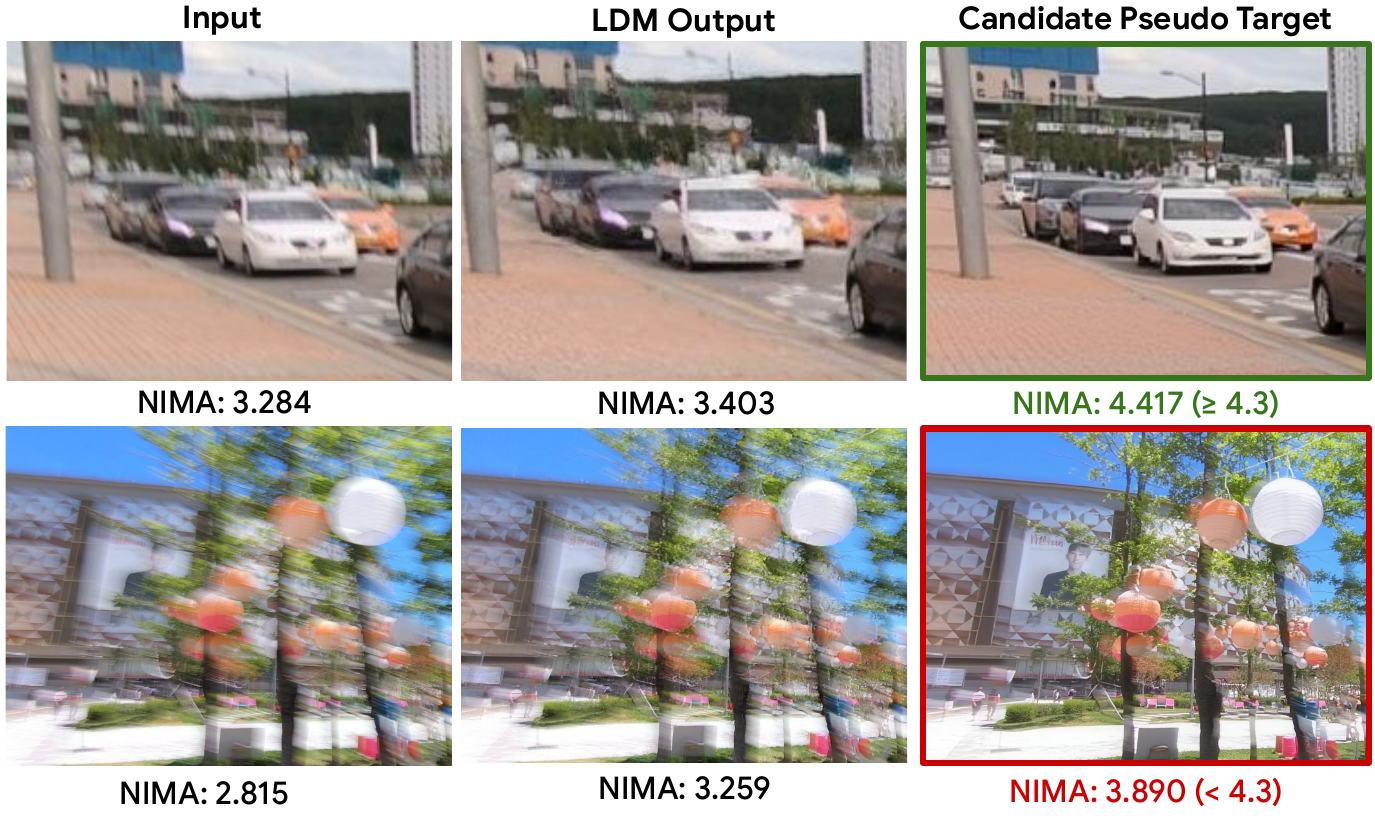}
    \caption{\textbf{Selection of targets based on image quality assessment on REDS training set.} Top row: A successful selection where the generated target scored above the NIMA threshold ($\ge 4.3$). Bottom row: A rejection case where the target quality is insufficient.}
    \label{fig:filter_visual_supp}
\end{figure}

Figure~\ref{fig:filter_visual_supp} visualizes this selection process. The top row demonstrates a successful restoration that preserves semantic integrity and passes the quality gate. Conversely, the bottom row illustrates a rejection case where the foundation model failed to recover valid structures. Table~\ref{tab:filter_stats} details the quantitative statistics of this process. Using a consistent NIMA threshold of $\alpha = 4.2$ across all datasets, we observe pass rates ranging from 69.1\% to 86.4\%. This variation reflects the complexity of different out-of-distribution domains. For instance, the REDS dataset, which features complex motion blur, exhibits the lowest pass rate (69.1\%), indicating that a significant portion of the initial restorations were too degraded to be reliable.

\vspace{4pt}
\noindent\textbf{Ablation on Generation Strategy.}
We analyze the impact of the generation strategy via a two-step evaluation: (1) assessing the intrinsic visual quality of the generated targets, and (2) evaluating the downstream performance of the adapted model. We compare our proposed pipeline—which utilizes Attention Injection proposed in RF-Solver~\cite{wang2025taming}—against two baselines: (1) the same inversion pipeline \textit{without} Attention Injection, and (2) standard SDEdit~\cite{sdedit}.

\begin{figure}[t]
    \centering
    \includegraphics[width=0.85\linewidth]{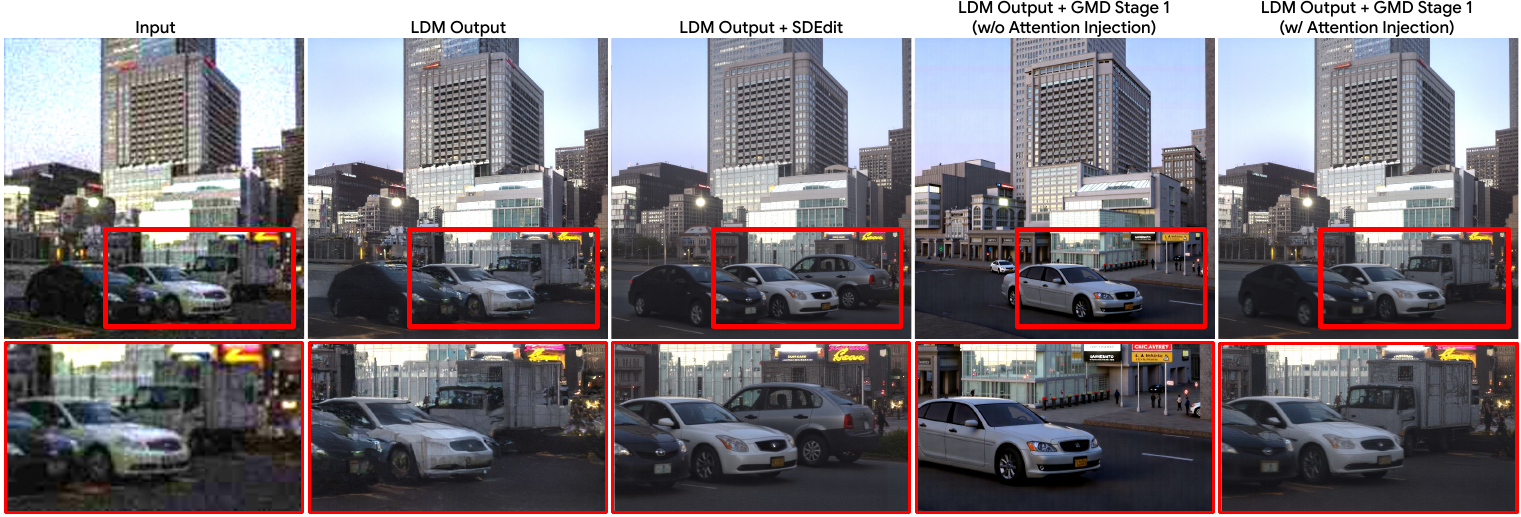}
    \caption{\textbf{Visual comparison of target generation strategies (Stage 1).} We evaluate the intrinsic quality of different generation methods. Standard baselines like SDEdit~\cite{sdedit} or the flow matching inversion without Attention Injection help remove artifacts but suffer from severe content drift and structural distortion (e.g., altering vehicle geometry). In contrast, our adopted method using \textbf{RF-Solver~\cite{wang2025taming}} (with Attention Injection) effectively restores details while strictly preserving the original structural layout. \textit{(Note: `GMD' refers to our GMD method.)}}
    \label{fig:pseudo_target_ablation_supp}
\end{figure}

\vspace{2pt}
\noindent\textit{1. Intrinsic Generation Quality.} We first examine the visual fidelity of the targets generated in Stage 1. Qualitative comparisons in Figure~\ref{fig:pseudo_target_ablation_supp} reveal that methods lacking Attention Injection (both SDEdit and the ``No Attention'' pipeline) struggle to maintain structural consistency; they often hallucinate new geometries or alter the semantic identity of objects. In contrast, RF-Solver utilizes attention keys and values from the forward process to guide the generation, ensuring strict structural fidelity. Given the visible structural failures of the ``No Attention'' variant, it is deemed unsuitable for supervising a restoration model, as it would bias the network toward learning geometric distortions.

\vspace{2pt}
\noindent\textit{2. Downstream Adaptation Performance.} Based on the visual analysis, we utilize the RF-Solver-generated targets for the downstream adaptation task. Table~\ref{tab:ablation_solver} presents the quantitative results, comparing our full GMD pipeline against the SDEdit-based adaptation. While SDEdit achieves a high MUSIQ score, it lags significantly in fidelity metrics (PSNR, SSIM, LPIPS). Our method significantly outperforms the SDEdit baseline across key fidelity and perceptual metrics (improving PSNR by 0.15 dB and FID by 2.45), confirming that the superior structural integrity provided by Attention Injection is crucial for effective domain adaptation.

\begin{table}[t]
\centering
\scriptsize
\setlength{\tabcolsep}{4pt}
\caption{\textbf{Ablation of Target Generation Method on REDS Dataset.} We compare downstream adaptation performance using different generation strategies. While SDEdit achieves competitive non-reference scores, it suffers from lower fidelity. Using RF-Solver (Inversion-based with Attention Injection) preserves structural integrity, leading to the best balance of fidelity (PSNR/SSIM) and perceptual quality (FID/LPIPS).}
\label{tab:ablation_solver}
\resizebox{\columnwidth}{!}{%
\begin{tabular}{l cccccc}
\toprule
\multirow{2}{*}{\textbf{Method}} & \multicolumn{5}{c}{\textbf{Perceptual Quality}} & \textbf{Distortion} \\
\cmidrule(lr){2-6} \cmidrule(lr){7-7}
 & \textbf{LPIPS}$\downarrow$ & \textbf{NIMA}$\uparrow$ & \textbf{MUSIQ}$\uparrow$ & \textbf{FID}$\downarrow$ & \textbf{CLIPIQA}$\uparrow$ & \textbf{PSNR / SSIM}$\uparrow$ \\
\midrule
Based on SDEdit~\cite{sdedit} & 0.180 & 4.419 & \best{64.05} & 34.09 & 0.375 & 24.20 / .675 \\
\rowcolor[HTML]{E8F4F8} \textbf{Based on RF-Solver (Ours)} & \best{0.179} & \best{4.460} & 63.67 & \best{31.64} & \best{0.404} & \best{24.35} / \best{.682} \\
\bottomrule
\end{tabular}%
}
\end{table}

\subsection{Generality and Additional Baseline Comparisons}
\label{sec:supp_generality}
We provide additional evaluation results to demonstrate the generality of GMD across different teacher and student capacities and compare GMD against more recent SOTA baselines.

\noindent\textbf{Model Generality (Teacher and Student Ablations).} 
To prove GMD is a generalizable framework rather than reliant on specific model capacities, we ablated both the teacher and student models:
\begin{itemize}
    \item \textbf{Different Students (SwinIR CNN student):} We applied GMD to a lightweight, non-diffusion CNN model (\textbf{SwinIR}, 11.8M parameters). SwinIR adapted with GMD (SwinIR+GMD) significantly improves the unadapted baseline FID from 38.40 to 35.93 at a very low latency of \textbf{0.20s} (Table~\ref{tab:supp_reds_baselines}).
    \item \textbf{Different Teachers (SD 1.3B teacher):} We ablated the massive 12B FLUX teacher with the $9.2\times$ smaller Stable Diffusion (\textbf{SD}, 1.3B) model during pseudo-target generation. Table~\ref{tab:supp_iphone_baselines} shows that the SD-adapted model still achieves substantial gains on the DPED-iPhone dataset, lowering NIQE to 4.768 compared to the baseline's 5.467. 
\end{itemize}
This confirms GMD extracts and aligns natural image priors regardless of the specific architectures used.

\noindent\textbf{Comparison against Fine-Tuned SOTA Baselines.}
We compared GMD against more recent SOTA restoration baselines: LucidFlux (FLUX-Based), HYPIR, and DiffBIR. Tables~\ref{tab:supp_reds_baselines} and~\ref{tab:supp_iphone_baselines} show that even massive 13.6B fine-tuned models suffer from severe out-of-distribution manifold drift under domain shifts, whereas GMD's offline manifold alignment systematically corrects this shift and comprehensively outperforms them.

\begin{table}[h]
\centering
\small
\setlength{\tabcolsep}{3.5pt}
\caption{\textbf{Motion Deblur on REDS Baselines \& Ablations (Supplement).}}
\label{tab:supp_reds_baselines}
\vspace{1.5mm}
\resizebox{\columnwidth}{!}{
\begin{tabular}{l|cc|cccc}
\toprule
\textbf{Method} & \textbf{Params} & \textbf{Lat.(s)} & \textbf{PSNR}$\uparrow$ & \textbf{SSIM}$\uparrow$ & \textbf{LPIPS}$\downarrow$ & \textbf{FID}$\downarrow$ \\ \midrule
\multicolumn{7}{l}{\textit{Additional SOTA Baselines}} \\ \midrule
LucidFlux (FLUX-Based) & $\sim$13.6B & 15.70 & 18.81 & 0.482 & 0.376 & 92.13 \\
HYPIR & $\sim$0.9B & 1.92 & 20.70 & 0.557 & 0.415 & 124.04 \\
DiffBIR & $\sim$0.9B & 4.25 & 22.61 & 0.574 & 0.329 & 67.22 \\ \midrule
\multicolumn{7}{l}{\textit{Target-Domain}} \\ \midrule
LDM-Deblur (Base) & 1.3B & 1.79 & 24.08 & 0.678 & 0.183 & 37.67 \\
LDM-Deblur+\textbf{GMD (FLUX)} & 1.3B & 1.79 & \textbf{24.35} & \textbf{0.682} & \textbf{0.179} & \textbf{31.64} \\ \midrule
\multicolumn{7}{l}{\textit{Student Agnostic}} \\ \midrule
SwinIR (Base) & 11.8M & \textbf{0.20} & 26.19 & 0.797 & 0.227 & 38.40 \\
SwinIR+\textbf{GMD (FLUX)} & 11.8M & \textbf{0.20} & \textbf{26.22} & \textbf{0.804} & \textbf{0.210} & \textbf{35.93} \\ \bottomrule
\end{tabular}
}
\end{table}

\begin{table}[h]
\centering
\small
\setlength{\tabcolsep}{3.5pt}
\caption{\textbf{Real-World SR on DPED-iPhone Baselines \& Ablations (Supplement).}}
\label{tab:supp_iphone_baselines}
\vspace{1.5mm}
\resizebox{\columnwidth}{!}{
\begin{tabular}{l|cc|cccc}
\toprule
\textbf{Method} & \textbf{Params} & \textbf{Lat.(s)} & \textbf{NIQE}$\downarrow$ & \textbf{BRISQUE$\downarrow$} & \textbf{MANIQA}$\uparrow$ & \textbf{MUSIQ}$\uparrow$ \\ \midrule
\multicolumn{7}{l}{\textit{Additional SOTA Baselines}} \\ \midrule
LucidFlux (FLUX-Based) & $\sim$13.6B & 10.30 & 5.446 & 30.278 & 0.639 & 62.71 \\
HYPIR & $\sim$0.9B & 1.24 & 6.200 & 18.970 & 0.626 & 56.96 \\
DiffBIR & $\sim$0.9B & 2.77 & 5.973 & 17.140 & 0.625 & 58.87 \\ \midrule
\multicolumn{7}{l}{\textit{Teacher Agnostic}} \\ \midrule
LDM-SR+\textbf{GMD (SD 1.3B)} & 1.3B & 1.17 & 4.768 & 16.981 & \textbf{0.632} & 59.00 \\ \midrule
\multicolumn{7}{l}{\textit{Target-Domain}} \\ \midrule
LDM-SR (Base) & 1.3B & 1.17 & 5.467 & 19.020 & 0.543 & 49.30 \\
LDM-SR+\textbf{GMD (FLUX)} & 1.3B & 1.17 & \textbf{4.300} & \textbf{16.350} & 0.627 & \textbf{59.45} \\ \bottomrule
\end{tabular}
}
\end{table}

\newpage
\subsection{Additional Visual Results}
\label{sec:supp_visuals}

\begin{figure}[t]
    \centering
    \includegraphics[width=\linewidth]{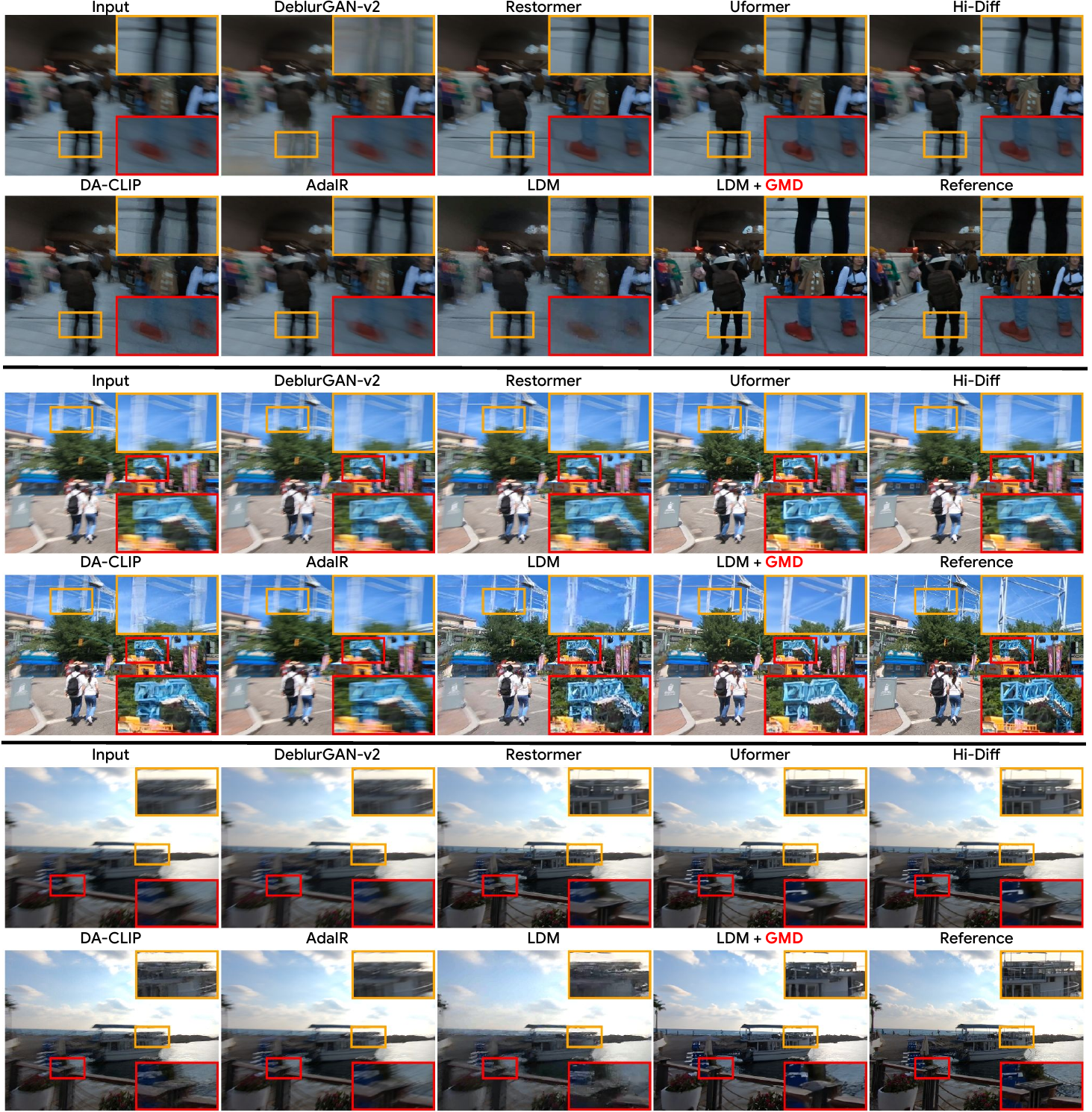}
    \caption{\textbf{Additional Qualitative comparison on the REDS deblurring dataset.} The GoPro-trained baseline (mismatched) leaves residual motion blur. GMD successfully adapts to the out-of-distribution domain, producing sharper, more detailed, and perceptually superior restorations. \textit{(Note: `LDM + GMD' refers to our GMD method.)}}
    \label{fig:deblur_visual_reds_supp}
\end{figure}

\begin{figure}[t]
    \centering
    \includegraphics[width=0.95\linewidth]{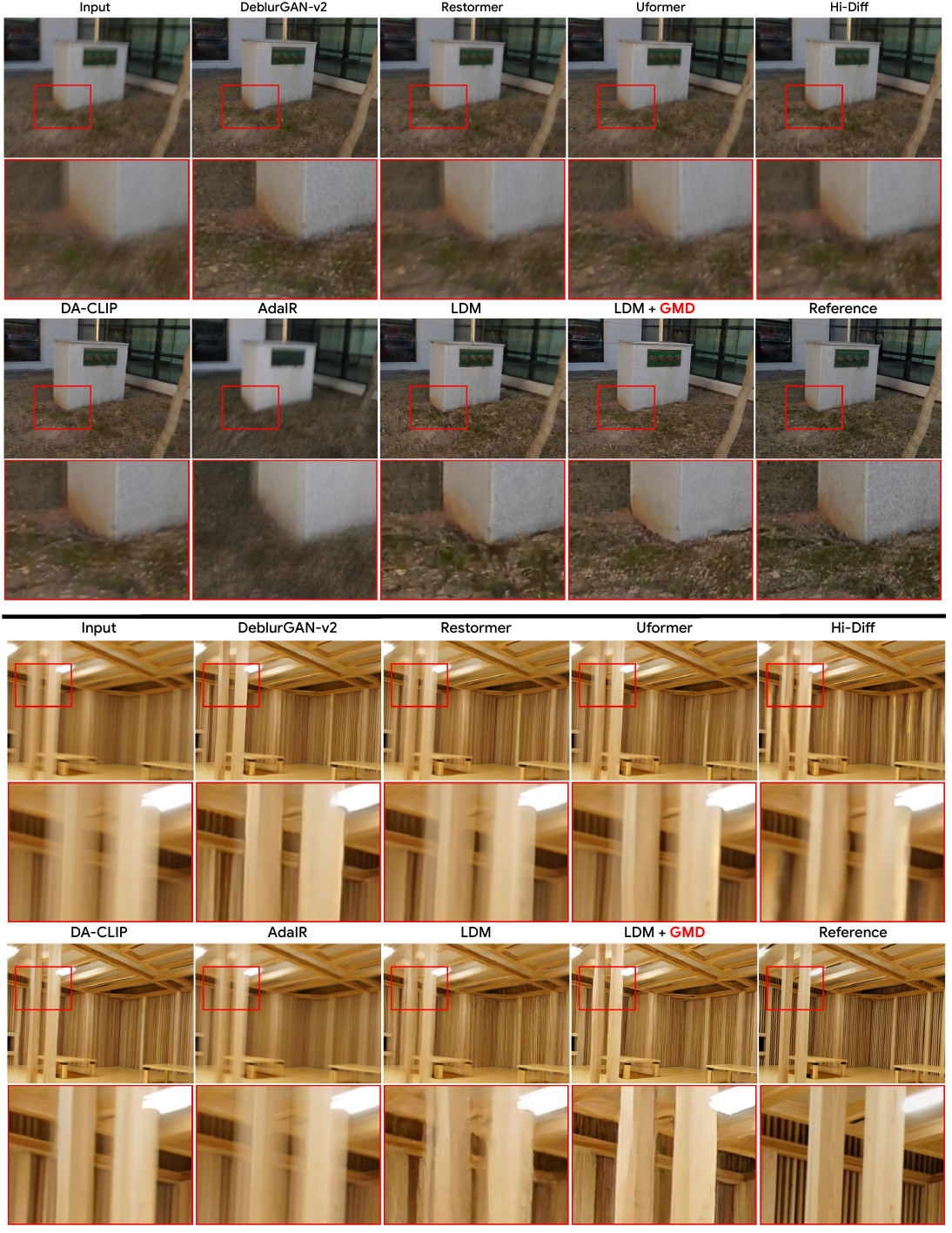}
    \caption{\textbf{Additional Qualitative comparison on the RealBlur-J deblurring dataset (Set 1).} Comparison showing the adaptation to real-world low-light blur. GMD recovers text and fine structures that are lost by the baseline model. \textit{(Note: `LDM + GMD' refers to our GMD method.)}}
    \label{fig:deblur_visual_realblurj_supp_1}
\end{figure}

\begin{figure}[t]
    \centering
    \includegraphics[width=0.95\linewidth]{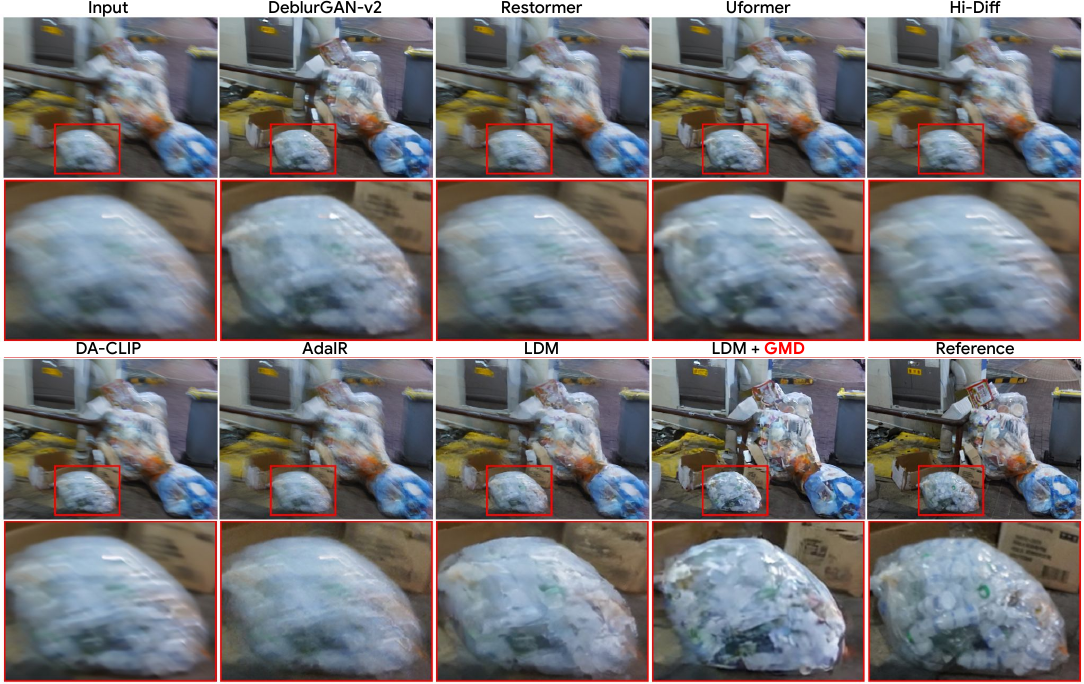}
    \caption{\textbf{Additional Qualitative comparison on the RealBlur-J deblurring dataset (Set 2).} GMD successfully adapts to the out-of-distribution domain, producing sharper, more detailed, and perceptually superior restorations. \textit{(Note: `LDM + GMD' refers to our GMD method.)}}
    \label{fig:deblur_visual_realblurj_supp_2}
\end{figure}

\begin{figure}[t]
    \centering
    \includegraphics[width=\linewidth]{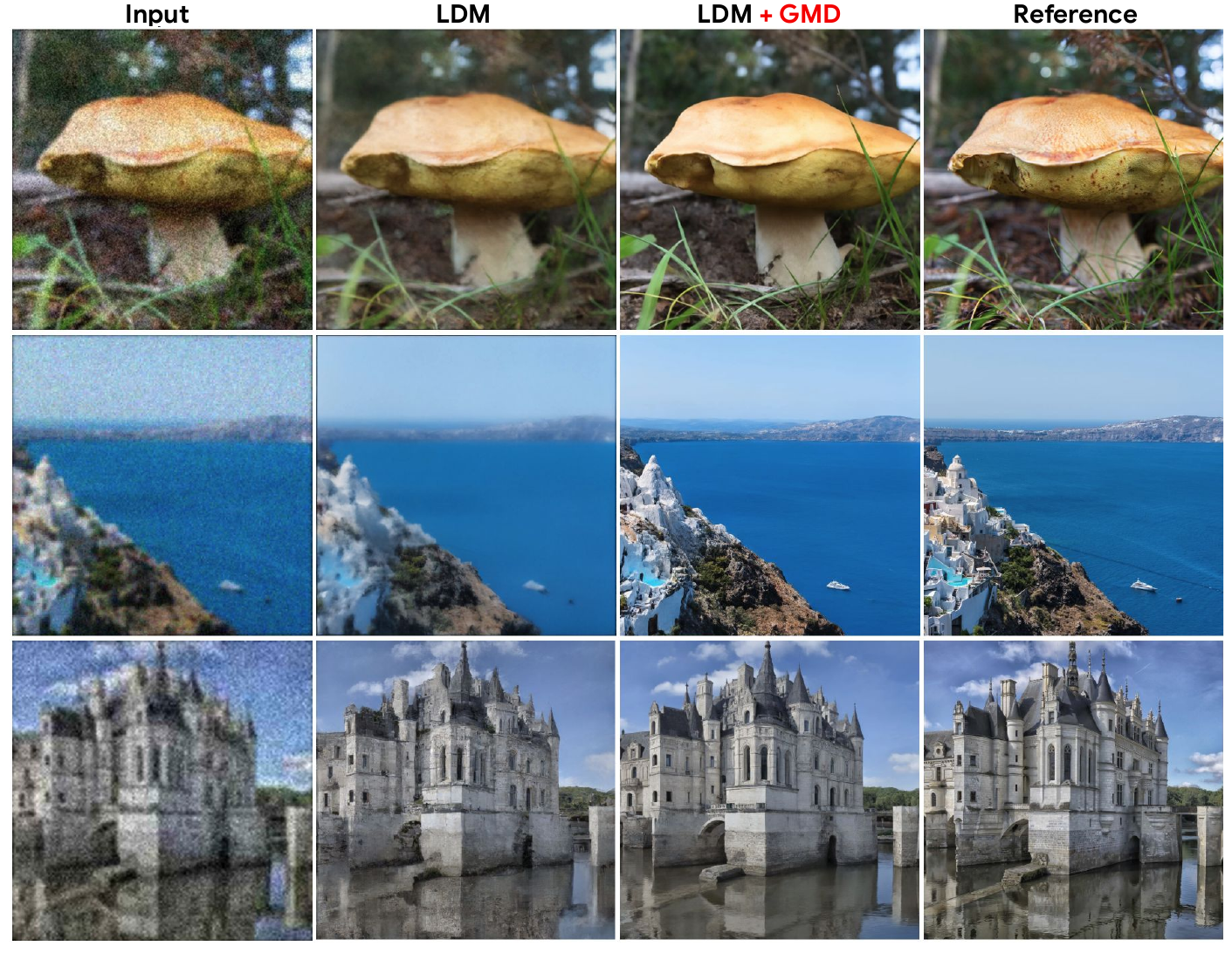}
    \caption{\textbf{Additional Qualitative comparison on Synthetic Super-Resolution (Weak $\to$ Strong).} The baseline model, pre-trained only on weak degradations, fails to generalize to the heavy noise and blur in the target domain. GMD successfully adapts to the stronger degradation profile, producing clean and sharp restorations. \textit{(Note: `LDM + GMD' refers to our GMD method.)}}
    \label{fig:sr_visual_div2k_supp_1}
\end{figure}

\begin{figure}[t]
    \centering
    \includegraphics[width=\linewidth]{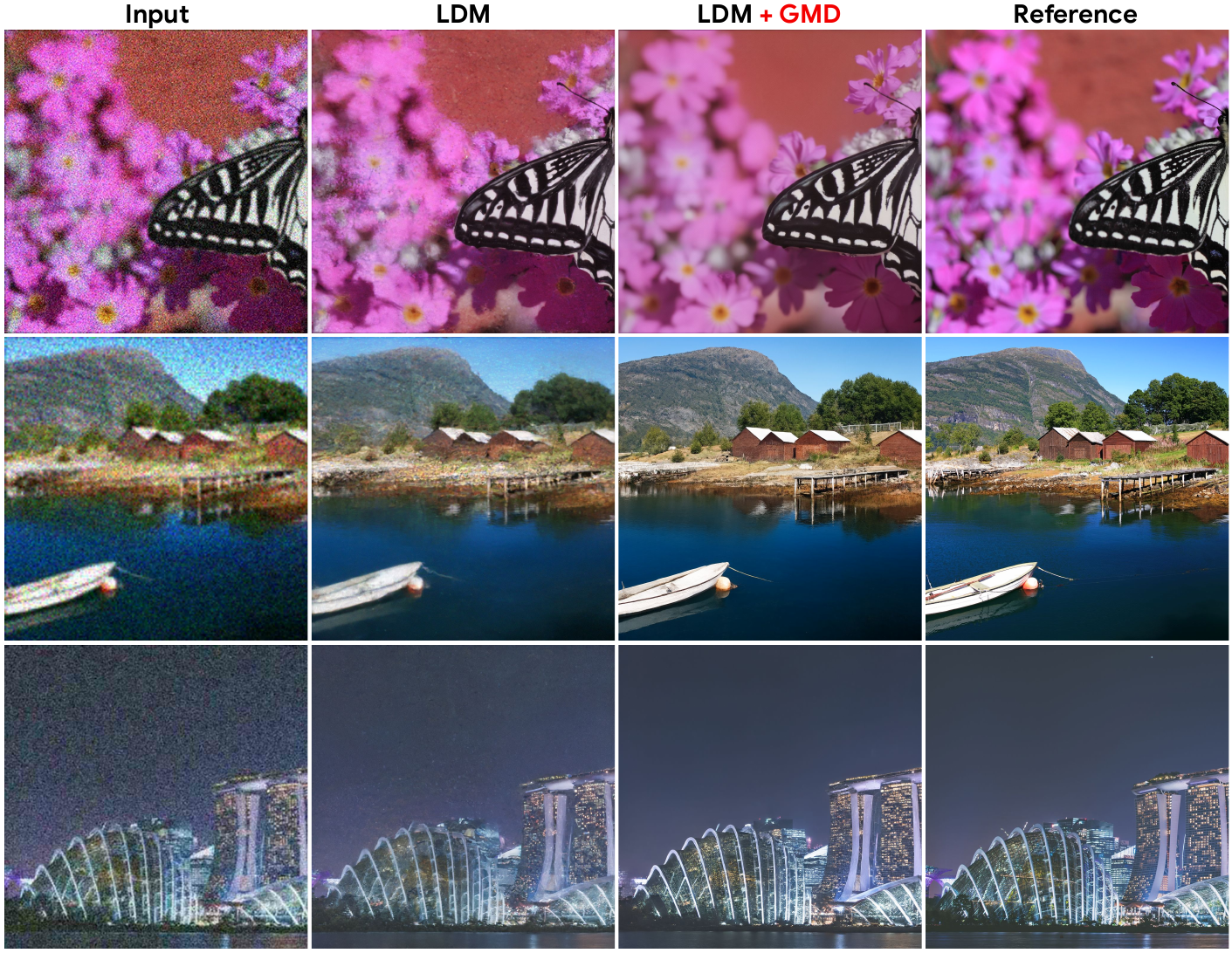}
    \caption{\textbf{Additional Qualitative comparison on Synthetic Super-Resolution (Weak $\to$ Strong).} GMD demonstrates robust adaptation to severe degradations, effectively removing noise and sharpening details without requiring paired ground truth. \textit{(Note: `LDM + GMD' refers to our GMD method.)}}
    \label{fig:sr_visual_div2k_supp_2}
\end{figure}

\begin{figure}[t]
    \centering
    \includegraphics[width=\linewidth]{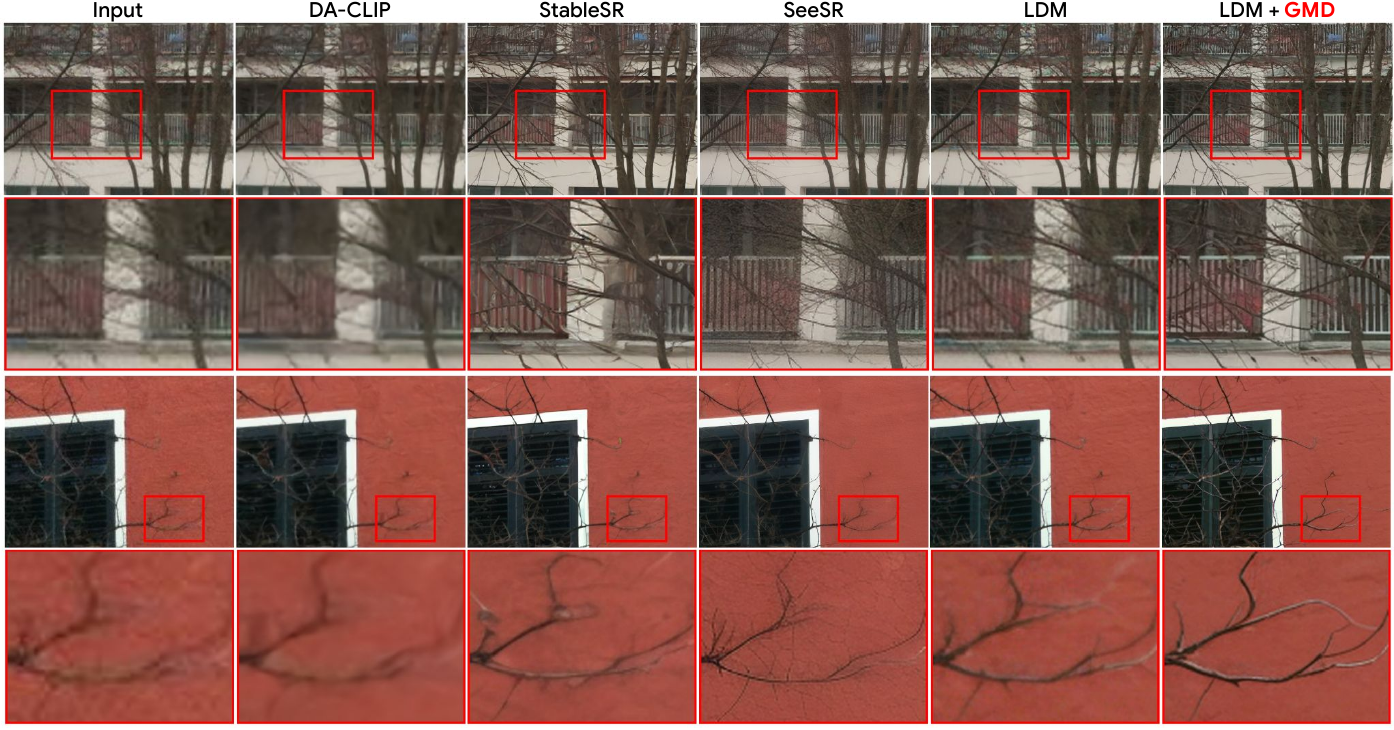}
    \caption{\textbf{Additional Qualitative comparison on Real-World Super-Resolution (Synthetic $\to$ DPED-iPhone).} The baseline model, pre-trained on synthetic data, fails to generalize to the complex sensor noise and compression artifacts of the iPhone camera. GMD successfully adapts to this real-world distribution, producing visually superior results. \textit{(Note: `LDM + GMD' refers to our GMD method.)}}
    \label{fig:sr_visual_iphone_supp_1}
\end{figure}

In this section, we provide a comprehensive qualitative evaluation to further demonstrate the efficacy of GMD in bridging the domain gap for image restoration. We extend the analysis from the main paper with additional comparisons across three distinct adaptation scenarios where paired ground truth is unavailable:

\begin{itemize}[leftmargin=1.5em, topsep=2pt, itemsep=2pt]
    \item \textbf{Deblurring (GoPro $\to$ REDS / RealBlur-J):} Figures~\ref{fig:deblur_visual_reds_supp}, \ref{fig:deblur_visual_realblurj_supp_1}, and~\ref{fig:deblur_visual_realblurj_supp_2} illustrate adaptation to complex video motion blur and real-world low-light motion blur settings.
    \item \textbf{Synthetic Super-Resolution (Weak $\to$ Strong):} Figures~\ref{fig:sr_visual_div2k_supp_1} and~\ref{fig:sr_visual_div2k_supp_2} show the model's ability to generalize to unseen, higher-intensity degradations.
    \item \textbf{Real-World Super-Resolution (Synthetic $\to$ DPED-iPhone):} Figure~\ref{fig:sr_visual_iphone_supp_1} demonstrates adaptation to real-world sensor noise and compression artifacts typical of mobile photography.
\end{itemize}

\noindent Across all settings, the unadapted baselines exhibit characteristic failures due to distribution shift—such as ringing, residual noise, or over-smoothing. In contrast, GMD successfully aligns with the target domain, recovering sharp high-frequency details and natural textures.

\end{document}